\documentclass[letterpaper, 10 pt, conference]{ieeeconf}  

\IEEEoverridecommandlockouts                              
\usepackage{times}

\usepackage{multicol}
\usepackage{graphics}
\usepackage[bookmarks=true]{hyperref}
\usepackage{amsmath,amssymb,bm}
\usepackage{setspace}
\usepackage[Symbol]{upgreek}
\usepackage{graphicx,subfigure}
\usepackage{graphics}
\usepackage{floatrow} 
\usepackage{multirow}
\usepackage{array}
\usepackage{enumerate}
\usepackage{sidecap}
\usepackage[all]{xy}
\usepackage{authblk}
\usepackage{comment}
\usepackage{changepage}   
\usepackage{wrapfig}
\usepackage{url}
\usepackage{tabularx,pbox,booktabs,caption}
\usepackage{flushend}
\usepackage{anyfontsize}
\usepackage{float}
\usepackage{todonotes}
\usepackage[percent]{overpic}
\graphicspath{{./pics/}}
\usepackage{amssymb}
\usepackage{dblfloatfix}    
\usepackage{booktabs}
\usepackage{amsfonts}
\usepackage[T1]{fontenc}

\usepackage{mathabx}
\usepackage{upgreek}
\usepackage{algpseudocode}
\usepackage{algorithm}

\newcommand{\RE}[1]{\textcolor{black}{{}#1}}
\newcommand{\Wording}[1]{\textcolor{black}{{}#1}}

\usepackage{amsthm}

\algnewcommand\algorithmicforeach{\textbf{for each}}
\DeclareMathOperator*{\argmax}{arg\,max}

\algdef{S}[FOR]{ForEach}[1]{\algorithmicforeach\ #1\ \algorithmicdo}
\algnewcommand\algorithmicinput{\textbf{Input:}}
\algnewcommand\INPUT{\item[\algorithmicinput]}
\algnewcommand\algorithmicoutput{\textbf{Output:}}
\algnewcommand\OUTPUT{\item[\algorithmicoutput]}

\title{
State-Continuity Approximation of Markov Decision Processes via Finite Element Methods
for Autonomous System Planning 
}

\author{Junhong Xu, Kai Yin, Lantao Liu
\thanks{J. Xu and L. Liu are with the Luddy School of Informatics, Computing, and Engineering  at Indiana University, Bloomington, IN 47408, USA. E-mail:
        {\tt\small \{xu14, lantao\}@iu.edu}.
K. Yin is with Expedia Group. E-mail:
        {\tt\small kyin@expediagroup.com}.   J.X. and K.Y. equally contributed. 
}
}

\begin{document}

\maketitle
\thispagestyle{empty}
\pagestyle{empty}

\begin{abstract}
Motion planning under uncertainty for an autonomous system can be formulated as
a Markov Decision Process with a continuous state space. 
In this paper, we propose a novel solution to this decision-theoretic planning problem that directly obtains the continuous value function with only the first and second moments of the transition probabilities, 
\Wording{alleviating the requirement for an explicit transition model in the literature.}
We achieve this by 
\Wording{expressing the value function as a linear combination of basis functions and approximating the Bellman equation by a partial differential equation,}
where the value function can be naturally constructed using a finite element method.
We have validated our approach via extensive simulations, and the evaluations reveal that \Wording{compared}
to baseline methods, our solution leads to 
\Wording{better planning results}
in terms of path smoothness, travel distance, and time costs.
\end{abstract}

\section{INTRODUCTION}




Many autonomous vehicles that operate in flow fields, e.g. aerial and marine vehicles, can be easily \Wording{influenced} by environmental disturbances. For example, autonomous marine vehicles might experience ocean currents as in Fig.~\ref{fig:dynamic_ocean}.
\Wording{Vehicles' uncertain} behavior in an externally disturbed environment can be modeled using a {decision-theoretic planning} framework where the substrate is the Markov Decision Process (MDP)~\cite{sutton2018reinforcement}.  
Since vehicles operate in a continuous domain, to obtain a highly accurate solution, it is desirable to solve the continuous-state-space MDP directly. 
However, this is generally difficult because the exact algorithmic solutions are known to be computationally challenging~\cite{bellman2015adaptive}. 
\Wording{Besides, finding the solutions to MDPs
typically requires knowing an accurate transition model,} which is usually an unrealistic assumption.
Existing works that apply the MDP \Wording{for} navigating aerial vehicles~\cite{al2013wind} or autonomous underwater vehicles (AUVs)~\cite{pereira2013risk, feyzabadi2014risk} typically use a simplified version of the continuous state space MDP.
They represent the original MDP using a grid-map based representation and assume the vehicle can only transit to adjacent \Wording{grid cells}. 
This coarse simplification 
\Wording{poorly}
characterize the original problem, and thus may lead to inconsistent solutions. 

In this work, we propose a novel method that obtains a high-quality solution in continuous state space MDPs without requiring the exact form of the transition function.
Compared to the majority of continuous MDP frameworks, we tackle the difficulties in large scale MDP problems from a different perspective: the integration of two layers of approximations. The value function is approximated by a 
\Wording{linear combination of basis functions}
and the Bellman equation is approximated by a diffusion type partial differential equation (PDE)\RE{, which allows us to express the Bellman equation using minimum characteristics of transition probabilities.} 
This combination naturally leads to the applications of the Finite Element \Wording{Method} (FEM) for the solution. 

Specifically, we first approximate the value function by a weighted linear combination of finite basis functions. 
Then \Wording{using} the Taylor expansion of the value function \cite{braverman2018taylor, buchli2017optimal}, we show that it satisfies a diffusion-type PDE, which only depends on the first and second moments of the transition probability.
We \Wording{apply the FEM} to solve the PDE with suitable boundary conditions. 
\RE{The method is based on discretization of the workspace into small patches in which the continuous value function may be approximated by a finite linear combination of basis functions, and the resulting approximation naturally extends over the entire continuous workspace.
}
Combining these procedures, we propose an approximate policy iteration algorithm to obtain the final policy and a continuous value function. Our framework in principle allows us to compute the policy on the entire planning domain (space). 
Finally, we validate our method \Wording{in a scenario involving navigating} a marine vehicle in the ocean. Our simulation results show that the proposed approach produces results superior to the classic grid-map based MDP solutions. 


{
\begin{figure}
    \centering
    \includegraphics[width=1.0\textwidth]{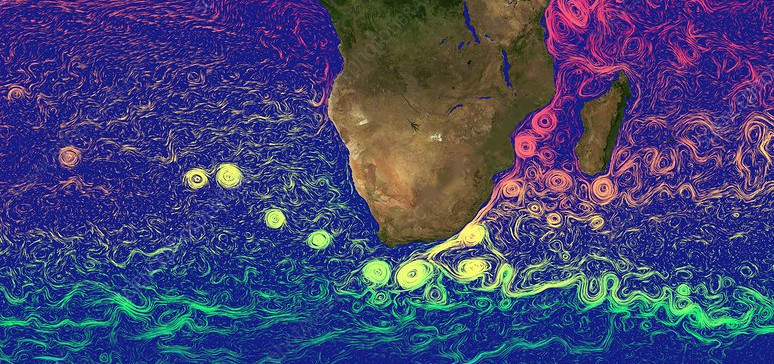}
    \caption{\small Oceans currents can cause significant disturbances for autonomous marine vehicles. 
    \Wording{The Agulhas Ring gyres of southern Africa are unusually strong.}
    (Source: NASA.) 
    }
    \label{fig:dynamic_ocean} 
\end{figure}
}


\section{RELATED WORK}
Planning under the presence of \Wording{action} uncertainty can be viewed as a decision-theoretic planning problem which is usually modeled \Wording{using MDPs~\cite{BoutilierDTP99}}. 
Most existing methods that solve MDPs in the robotic planning literature use tabular methods, i.e., converting the continuous state space into a set of discrete indices~\cite{ thrun2002probabilistic, sutton2018reinforcement}. 
For example, in~\cite{al2013wind}, the authors model the task of minimum energy-based planning for aerial vehicles under {a} wind disturbance using a finite-state MDP. 
Similar work~\cite{huynh2015predictive, liu2018solution, kularatne2018optimal, Liu-RSS-19} addresses tasks of minimum time and energy planning under uncertain ocean currents as an MDP problem. 
\RE{
These grid-based solutions typically use a histogram-type approximation and assume the same values within each discretized state representation.
To achieve good accuracy, the histogram-type approximation usually requires fine discretization resolution, which leads to significant computational time.
In contrast, our method approximates the value function in a continuous form directly by the basis function. 
}

Another well-known approach for optimal planning is based on incremental sampling-based methods, e.g., PRM* and RRT*~\cite{karaman2011sampling}.
These methods have been used in target searching~\cite{hollinger2015long}, ship hull inspection~\cite{hover2012advanced}, and autonomous driving~\cite{gonzalez2015review}. 
The \Wording{asymptotically} optimal behavior of these algorithms is the result of the principle of dynamic programming~\cite{arslan2015dynamic, arslan2016incremental}, which shares the same root as solving the \Wording{MDP problems}.
However, these algorithms are not suitable for planning under uncertainty problems because they assume the motion of the vehicle is deterministic. 
Recent work~\cite{huynh2012incremental, agha2014firm} extends the original RRT* and PRM* algorithms to \Wording{solve planning tasks} which have stochastic action effects. 
It builds a sequence of MDPs by sampling the states using {an} RRT-like method then performs asynchronous value iteration~\cite{bertsekas2011dynamic} iteratively.
The above incremental sampling-based methods find trajectories and attempt to follow them.  
In contrast to these methods, our proposed method finds the optimal value function directly, which in turn generates optimal behavior.  

It is also worth mentioning that many direct value function approximation methods have been proposed in machine learning and reinforcement learning~\cite{sutton2018reinforcement}.
Generally, the approximate value function is represented as a parametric functional form with weights~\cite{konidaris2011value, kober2013reinforcement}.
Fitted value iteration (FVI) is \Wording{one popular sampling method}
that approximates continuous value functions \cite{szepesvari2005finite, antos2008fitted}.
It requires batches of samples to approximate the true value function via regression methods. 
However, the behaviour of the algorithm is not well understood, and without \Wording{careful choice of}
a function approximator, the algorithm may diverge or become very \Wording{slow to converge}
~\cite{lizotte2011convergent,baird1995residual}.
In contrast to the sampling methods, we attempt to approximate the value function using 
second-order Taylor expansion and calculate the resulting partial differential equation via a finite element method 
\cite{babuvska1971error}.

\section{Problem Description and Formulation}
This section \Wording{describes decision-theoretic} planning via Markov Decision Processes (MDPs).
We consider planning of minimum time cost under uncertain actions caused by disturbances (\Wording{e.g., ocean currents, air turbulence}). 
This problem is usually formulated as an infinite time-horizon MDP.

We represent the infinite discrete time horizon MDP as a 4-tuple  $(\mathbb{S}, \mathbb{A}, \mathcal{T}, R)$. The continuous \textit{spatial} state space $\mathbb{S}$ denotes the entire autonomous vehicle planning region (or domain).
Accordingly, a state $s\in\mathbb{S}$ is a spatial point $(x, y)\in \mathbb{R}^2$ on the plane, and it indicates the location of the vehicle. 
When the autonomous vehicle starts from a state, it takes an action to move toward {the} next state. We model the action space $\mathbb{A}$ as a finite space.
An action depends on the state; that is, for each state $s\in\mathbb{S}$, we have a feasible action set $A(s)\subset\mathbb{A}$. 
The entire set of feasible state-action \Wording{tuples} is $\mathbb{F} := \{(s, a)\in \mathbb{S}\times\mathbb{A}\}$. 
There is a probability transition law $\mathcal{T}(s, a; \cdot)$ on $\mathbb{S}$ for all $(s, a)\in \mathbb{F}$. $\mathcal{T}(s, a; s')$ specifies the probability of transitioning to the state $s'$ given the current state $s$ with the \Wording{chosen} action $a$ constrained by system dynamics. 
The final element $R: \mathbb{F}\rightarrow\mathbb{R}^1$ is a real-valued reward function that depends on state and action.

We consider the class of deterministic Markov policies~\cite{puterman2014markov}, denoted by $\Pi$, i.e., the mapping $\pi: \mathbb{S}\rightarrow\mathbb{A}$ depends on the current state and the current time, and $\pi(s)\in A(s)$ for a given state. 
For a given initial state $s_0$, the expected discounted total reward is represented as:
\begin{equation}
    v^\pi(s_0)=\mathbb{E}^\pi_{s_0}\left[\sum_{k=0}^{\infty} \gamma^kR(s_k, a_k)\right],
    \label{MDP-value}
\end{equation}
where $\gamma\in[0, 1)$ is {a} discount factor that discounts the reward at a \Wording{geometrically} decaying rate. 
The aim is to find a policy $\pi^*$ to maximize the expected cumulative discounted reward starting from the initial state $s_0$, i.e.,
\begin{equation}
    \pi^*(s_0)=\argmax_{\pi\in\Pi} v^\pi(s_0).
    \label{optimal-policy}
\end{equation}
Accordingly, the optimal value is denoted by $v^*(s_0)$.

Under certain conditions \cite{puterman2014markov, bertsekas1995dynamic}, it is well-known that the optimal \Wording{solution satisfies} the following recursive relationship
\begin{equation}
    v(s)=\max_{a\in A(s)}\left\{R(s, a) + \gamma\cdot\mathbb{E}^a[v(s')\mid s]\right\},
    \label{MDP-Bellman}
\end{equation}
where $\mathbb{E}^a[v(s')|s]=\int \mathcal{T}(s, a; s')\,v(s')ds'$. 
This is the \RE{Bellman optimality equation}, which serves as a basis to solve the problem \Wording{expressed by} Eq.~\eqref{MDP-value} and Eq.~\eqref{optimal-policy}. 
Popular algorithms to obtain the solutions include policy and value iteration algorithms as well as linear program based methods \cite{puterman2014markov}. 

\section{Methodology}
Our proposed methodological framework consists of four key interconnected elements for the solution to the MDP problem with a continuous or large-scale state space.
First, we approximate the value function by a \Wording{weighted linear combination} of basis functions. 
Once the weights are determined, the value function for the entire state space is readily \Wording{evaluated}. 
Secondly, to alleviate the \Wording{requirement for} an explicit transition model, 
we approximate the Bellman equation by a diffusion-type partial differential equation (PDE). 
This is achieved through Taylor expansion of the value function with respect to the state variable \cite{braverman2018taylor, buchli2017optimal}. 
The obtained PDE only requires the first and second moments of the transition probabilities, potentially leading to wide \Wording{applicability}. 
Thirdly, integrating the above key steps together naturally leads to the Finite Element {Method} (FEM) for the solution to value function approximation given a policy. 
The FEM framework allows us to transfer the approximate PDE to a linear system of equations whose solutions are exactly the weights in the linear combination of basis functions. 
Moreover, the resulting linear system depends on \Wording{finitely many} discrete states only. 
Finally, we propose a policy iteration algorithm based on the diffusion-type PDE and FEM to obtain the final policy and continuous approximate value function. 

\subsection{Value Function Approximation by Basis Functions}\label{sec:value-fn-approximation}
The value function plays a central role in the MDP. 
Because \Wording{daunting computational requirements} typically prevent any direct value computation on continuous or large scale states, value function approximation becomes necessary to obtain an approximate solution. 
We approximate the value function \Wording{as the linear span of}
a finite set of basis functions $\{\phi_i(s)\}$,
\begin{equation}\label{basis_func_appro}
    v^{\pi}(s) = \sum_{i=1}^n w_i^{\pi}\phi_i(s),
\end{equation}
where $w_i^{\pi}$ are weights under the policy $\pi$. Then the methodological focus shifts to tractable and efficient algorithms for the weight computation as well as the policy improvement. Basis functions 
\Wording{appropriate}
for the robotic motion planning will be specified within the following framework.  

\subsection{Diffusion-Type Approximation to Bellman Equation}\label{sec:Taylor-Value}

Besides the difficulties 
\Wording{in handling}
large scale continuous states, a critical feature of the Bellman optimality equation Eq.~\eqref{MDP-Bellman} is the requirement of \Wording{being able to evaluate exact transition probabilities to all the next states for all state-action pairs.}
Such requirement is unrealistic in many applications. Therefore, we seek an approximation to Eq.~\eqref{MDP-Bellman} that only needs the minimum characteristics of transition probabilities. 

We 
\Wording{subtract $v(s)$ from both sides of Eq.~\eqref{MDP-Bellman}}
and then take Taylor expansions of \Wording{the} value function around $s$ up to the second order \cite{braverman2018taylor}: 
\begin{align}
    0&=\max_{a\in A(s)}\Big\{R(s, a) + \gamma\,\left(\mathbb{E}^a[v(s')\mid s] - v(s)\right)\\ 
    &- (1-\gamma)\,v(s)\Big\} \nonumber\\
    &\Wording{\approx} \max_{a\in A(s)}\Big\{R(s, a) + \gamma\,\Big((\mu_s^a)^T\nabla v(s) +\frac{1}{2}\nabla\cdot \sigma_s^a\nabla v(s)\Big) \nonumber\\ 
    & - (1-\gamma)\,v(s)\Big\}, \label{bellman-pde}
\end{align}
where 
\begin{subequations}
\begin{align}\label{moments-eqn}
    \mu_s^a &= \int \mathcal{T}(s, a; s')(s' - s) ds', \\
    \sigma_s^a &= \int \mathcal{T}(s, a; s')(s' - s)(s' - s)^T ds'.
\end{align}
\end{subequations}
\Wording{The notation} $\cdot$ in Eq.~\eqref{bellman-pde} denotes the inner product. 
Here we assume that $v(s)$ \Wording{is continuously differentiable up to the second order.}
\Wording{For most} of motion planning problems where location may be modeled as state, we have $s=[x,y]^T$, the operator $\nabla \overset{\Delta}{=} [\partial/\partial x, \partial/\partial y]^T$, and 
$\nabla\cdot \sigma_s\nabla \overset{\Delta}{=}\sigma^{a}_{xx}\frac{\partial^2}{\partial x^2} 
    +2\sigma^{a}_{yx}\frac{\partial^2}{\partial y\partial x}
    +\sigma^{a}_{yy}\frac{\partial^2}{\partial y^2}$. 

Eq.~\eqref{bellman-pde} approximates the Bellman equation of the original problem. 
As a result, the solution approximates the optimal solution to the MDP. 
The benefit of \Wording{such an approximation} is that it only uses the first and second moments of the transition probabilities (Eq.~\eqref{moments-eqn}) rather than the full \Wording{expression.} 
 
Under the optimal policy, Eq.~\eqref{bellman-pde} is a diffusion type partial differential equation (PDE) with the optimal value function as its solution. Similarly, given a policy, the solution to the PDE from Eq.~\eqref{bellman-pde} is the associated value function. Because the obtained value function from a given policy can be used to improve the policy, 
it inspires us to \Wording{derive}
a policy iteration algorithm for the solution \cite{braverman2018taylor}: 
\begin{itemize}
\item In the policy evaluation stage, we solve for a diffusion type PDE with proper boundary conditions to obtain the value function $v(s)$; 
\item In the policy improvement stage, we search for a policy that maximizes the values of the right hand side of Eq. (\ref{bellman-pde}) with $v(s)$ obtained from the previous policy evaluation stage. 
\end{itemize}
The next section provides the PDE for the policy evaluation stage. 

\subsection{Partial Differential Equation Representation}\label{sec:PDE-Value}

It is well known that the suitable boundary conditions \Wording{must be}
imposed in order to obtain the appropriate solution to a PDE. 
We thus find such boundary conditions for the PDE from Eq.~\eqref{bellman-pde} and provide the policy evaluation approach. 

Since the value function does not have values outside the motion planning workspace, the directional derivative of the value function with respect to the \Wording{unit normal vector} at the boundary of the planning workspace must be zero. In addition, we constrain the value function at the goal state to be \Wording{a constant value} to ensure that there is a unique solution.
Let $\mathbb{S}$ in the MDP formulation be the entire continuous spatial planning region (called the domain), \Wording{denote} its boundary by $\partial\mathbb{S}$, and the goal state by $s_g$. 
We also use $\hat{\bm{n}}$ to denote the unit vector normal to $\partial\mathbb{S}$ pointing outward. Under the policy $\pi$, we aim to solve the following diffusion type PDE:
\begin{equation}
 -R(s, \pi(s)) =\gamma\,\Big((\mu_s^\pi)^T\nabla v(s) +\frac{1}{2}\nabla\cdot \sigma_s^\pi\nabla v(s)\Big) - (1-\gamma)\,v(s), 
 \label{diffusion-pde}
\end{equation}
with boundary conditions
\begin{subequations}\label{boundary-condition}
\begin{eqnarray}
    \RE{\sigma_s^\pi}\,\nabla v(s)\cdot \hat{\bm{n}} &=& 0, \mbox{  on } \partial\mathbb{S} \label{boundary-condition-a}\\
    v(s_g) &=& \RE{v_g},  \label{boundary-condition-b}
\end{eqnarray}
\end{subequations}
where $\mu_s^{\pi}$ and $\sigma_s^\pi$ indicate that $\mu_s$ and $\sigma_s$ are obtained under the policy $\pi$; \RE{$v_g$ is the value at the goal state.} The condition (\ref{boundary-condition-a}) is a type of homogeneous Neumann condition, and (\ref{boundary-condition-b}) can be thought of as a Dirichlet condition~\cite{Evens2010}.
We assume that the solution to the above Eq.~\eqref{diffusion-pde} with boundary conditions~\eqref{boundary-condition} exists. Other conditions for a well-posed PDE can be found in \cite{Evens2010}.

It is generally impossible to \Wording{obtain closed-form} solution to Eq.~\eqref{diffusion-pde}. One must resort to certain numerical methods. Next we will introduce a finite element method to construct the solution. The method just happens to leverage the linear approximation to the value function introduced in Section \ref{sec:value-fn-approximation}, and perfectly serves our objectives. 

\subsection{A Finite Element Method}\label{sec: fem_ped}

\begin{figure}[t]
    \centering
    \subfigure[Grids of States]
        {\label{fig:states-grids}\includegraphics[width=0.32\textwidth]{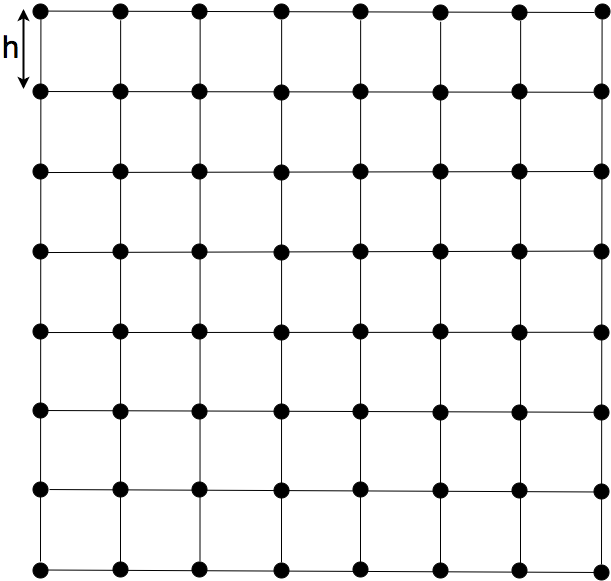}}
    \subfigure[Meshes]
        {\label{fig:meshes}\includegraphics[width=0.305\textwidth]{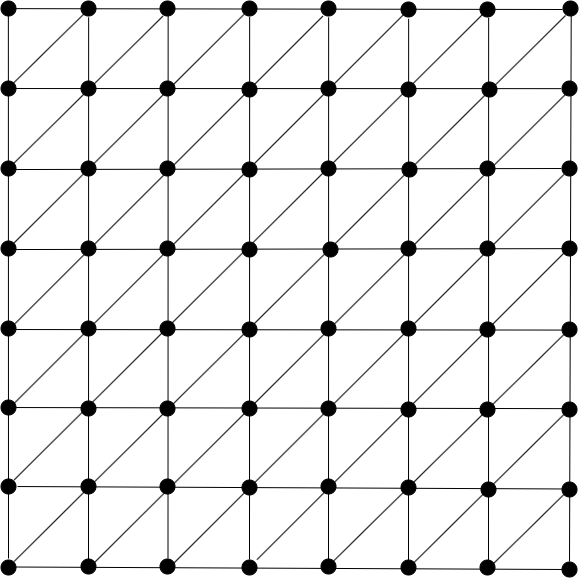}}
    \subfigure[Meshes on a subset of states]
        {\label{fig:meshes-subset}\includegraphics[width=0.304\textwidth]{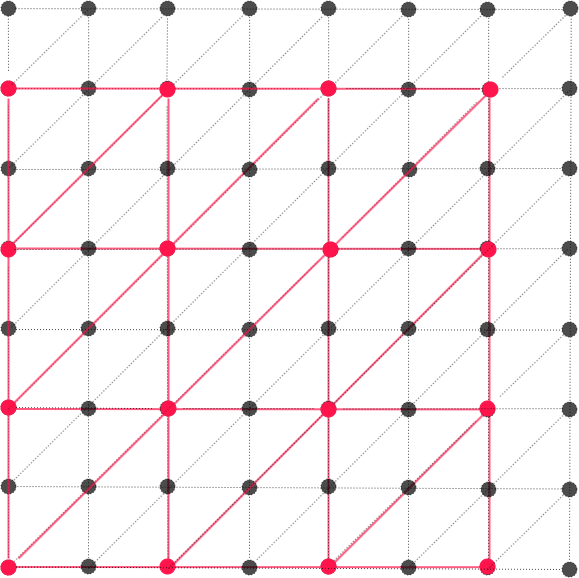}}
    \vspace{-10pt}
\caption{\small{Examples of \Wording{triangular} meshes with discrete states $\mathbb{S}'$.}}
\label{fig:finite-meshes} 
\end{figure}

\begin{figure}
\vspace{1pt}
\end{figure}
\begin{algorithm} [t]
\setstretch{1.0}
    \caption{Policy Iteration with State-Continuity Approximation and Finite Element Methods}\label{alg: API-FEM} 
  \begin{algorithmic}[1]
     \INPUT{Transition function $\mathcal{T}$, reward function $R$, discount $\gamma$, continuous spatial states $\mathbb{S}$, a finite subset $\mathbb{S'}
      \subseteq\mathbb{S}$, and goal state $s_g$.}
    \OUTPUT{Policy $\pi$ and continuous value function $v(s)$.}
    \State Initialize $\pi, \forall s \in \mathbb{S}$, and set $i=0$.
    \Repeat
      \State \parbox[t]{1.0\linewidth}{Step 1. Finite Element Based Policy evaluation.} 
      \State \parbox[t]{1.0\linewidth}{Compute $\mu_s^{\pi}$ and $\sigma_s^{\pi}$ on the finite states $\mathbb{S'}
      \subseteq\mathbb{S}$;}\ 
      
      \State \parbox[t]{.9\linewidth}{Obtain continuous value function $v^i$ by solving Eq.~\eqref{diffusion-pde} with boundary conditions Eq.~\eqref{boundary-condition}
      using Finite Element Method described in Section \ref{sec: fem_ped};\\} 
      
      \State Step 2. Policy improvement.
      \State \parbox[t]{0.9\linewidth}{Update the policy $\pi$ on $\mathbb{S}'$ using the value function $v^i$ according to the following equation:\\
            {\small$\pi(s) = \argmax_{a\in A(s)}\Big\{R(s, a) + \gamma\,\Big(\mu_s^T\,\nabla v^i(s) +\frac{1}{2}\nabla\cdot \sigma_s\nabla v^i(s)\Big) - (1-\gamma)\,v^i(s)$}
       } 
      \State \parbox[t]{.8\linewidth}{Set $i := i + 1$.} 
    \Until {$\pi(s)$ does not change for all $s \in \mathbb{S}'$}
  \end{algorithmic}
\end{algorithm}

Given a policy, we use a finite element method, 
\Wording{in particular, a Galerkin method,}
to compute the weights for the linear value function approximation Eq.~\eqref{basis_func_appro} through the PDE Eq.~\eqref{diffusion-pde} with boundary conditions Eq.~\eqref{boundary-condition}, and thereby \Wording{obtain} the value function approximation in the policy evaluation step.    

We will sketch the main ideas involved in finite element methods, and refer the readers to 
literature \cite{hughes2012finite, oden2012introduction} for a full account of the theory\footnote{In Appendix~\ref{appendix:fem}, we also provide an example of FEMs through a diffusion equation.}. 
The method consists of dividing the domain $\mathbb{S}$ into a finite number of simple subdomains, the finite elements, and then using \Wording{a variational form} (also called weak form) of the differential equation to construct the solution through the linear combination of basis functions over \Wording{elements}~\cite{oden2012introduction}.

The variational or weak form of the problem amounts to multiplying the both sides of Eq. (\ref{diffusion-pde}) by a test function $\omega(s)$ with suitable properties, and using integration-by-parts and 
the boundary condition Eq.~\eqref{boundary-condition} we have
{\small
\begin{align}
    -\int R\,\omega\,ds &=\gamma\,\int (\mu_s^\pi)^T\nabla v\,\omega\,ds -\frac{\gamma}{2}\int \sigma_s^\pi\nabla v\cdot\nabla \omega\,ds \nonumber\\
    &- (1-\gamma)\,\int v\,\omega\,ds. \label{weak-diffusion-pde}
\end{align}
}
Here the test function $w(s)=0$ on $s_g$. It can be shown the solution to this variational form is also the solution to the original form. 

\Wording{Next,} we partition the continuous domain $\mathbb{S}$
into suitable discrete elements. 
Typically \Wording{the mesh is constructed from triangular elements}, 
because they are general enough to handle all kinds of domain geometry. 
Suppose that the \Wording{elements} in our problem setting are 
\Wording{pinned to}
a finite set of discrete states $\mathbb{S}'\subset\mathbb{S}$. 
For example, Fig. \ref{fig:states-grids} shows a 
\Wording{continuous, square domain}
\Wording{approximated by an $8 \times 8$ set of discrete points $\mathbb{S}'$ separated by a constant distance $h$, where}
$h$ stands for the resolution parameter and the dark dots are the discrete states.
Fig. \ref{fig:meshes} shows 
\Wording{an example of triangular finite elements, where each vertex of each element corresponds to a state in $\mathbb{S}'$.}

In addition to the linear approximation to the value function Eq.~\eqref{basis_func_appro}, we also represent the test function by a linear combination of basis functions, i.e., $\omega(s)=\sum_{i=1}^n c_i\phi_i(s)$. 
We use the Lagrange interpolation polynomials as basis functions \Wording{for the test and value functions, constructed based on nodes of the triangle elements.}
Substituting the approximations of value and test functions into Eq.~\eqref{weak-diffusion-pde}, we get an equation for the weights $c_i$ and $w_i$.
Because coefficients $c_i$ should be arbitrary, this along with condition \Wording{(\ref{boundary-condition-b})} leads to a coupled system 
of linear algebraic equation of type $KW=F$. Each entry of the matrix $K$ corresponds to 
\Wording{integrals over the product of derivatives of basis functions}
on the right-hand side of Eq. (\ref{weak-diffusion-pde}). 
If we choose $n$ number of 
Lagrange interpolation polynomials, the $K$ is of $n\times n$ size. 
$F$ corresponds to \Wording{the remaining integrals} 
on the both hand sides of Eq. (\ref{weak-diffusion-pde}). Solving this linear system gives the estimates of $w_i$, i.e., the approximate value function $v(s)$.

We make two notes here. First, we may choose a relatively small number of state points to form $\mathbb{S}'$ for the finite element method. 
This is because the finite element method approximate the solution with high precision. 
Fig. \ref{fig:meshes-subset} provides an example of using 
a smaller number of \Wording{larger elements, which are pinned to fewer discrete states compared to Fig.~\ref{fig:meshes}.} 
The red dots 
\Wording{are the selected states and red triangles are the corresponding elements.}
We will demonstrate this point in numerical examples in Section~\ref{sect:expt}. 
Second, different types of applications and equations may require different 
\Wording{mesh designs.}

\subsection{Approximate Policy Iteration Algorithm}\label{sec: policy_iter}

We summarize our approximate policy iteration algorithm in Algorithm \ref{alg: API-FEM}.

In the policy evaluation step, we 
\Wording{apply the FEM to}
a subset $\mathbb{S'}\subset\mathbb{S}$. If the computational cost is a concern, we can further reduce the size of $\mathbb{S'}$ 
\Wording{by adjusting the resolution parameter $h$ used to create the subset $\mathbb{S}'$ in the preceding section.}
Note that the obtained value function is continuous on the entire planning region. Therefore, it is possible to evaluate policy $\pi$ on the whole state space $\mathbb{S}$.  




\section{Experiments}
\label{sect:expt}
\begin{figure}[t] 
    \vspace{2pt}
    \centering
    \subfigure[Classic Policy Iteration]
        {\label{fig:exact_weak_traj}\includegraphics[width=0.325\textwidth]{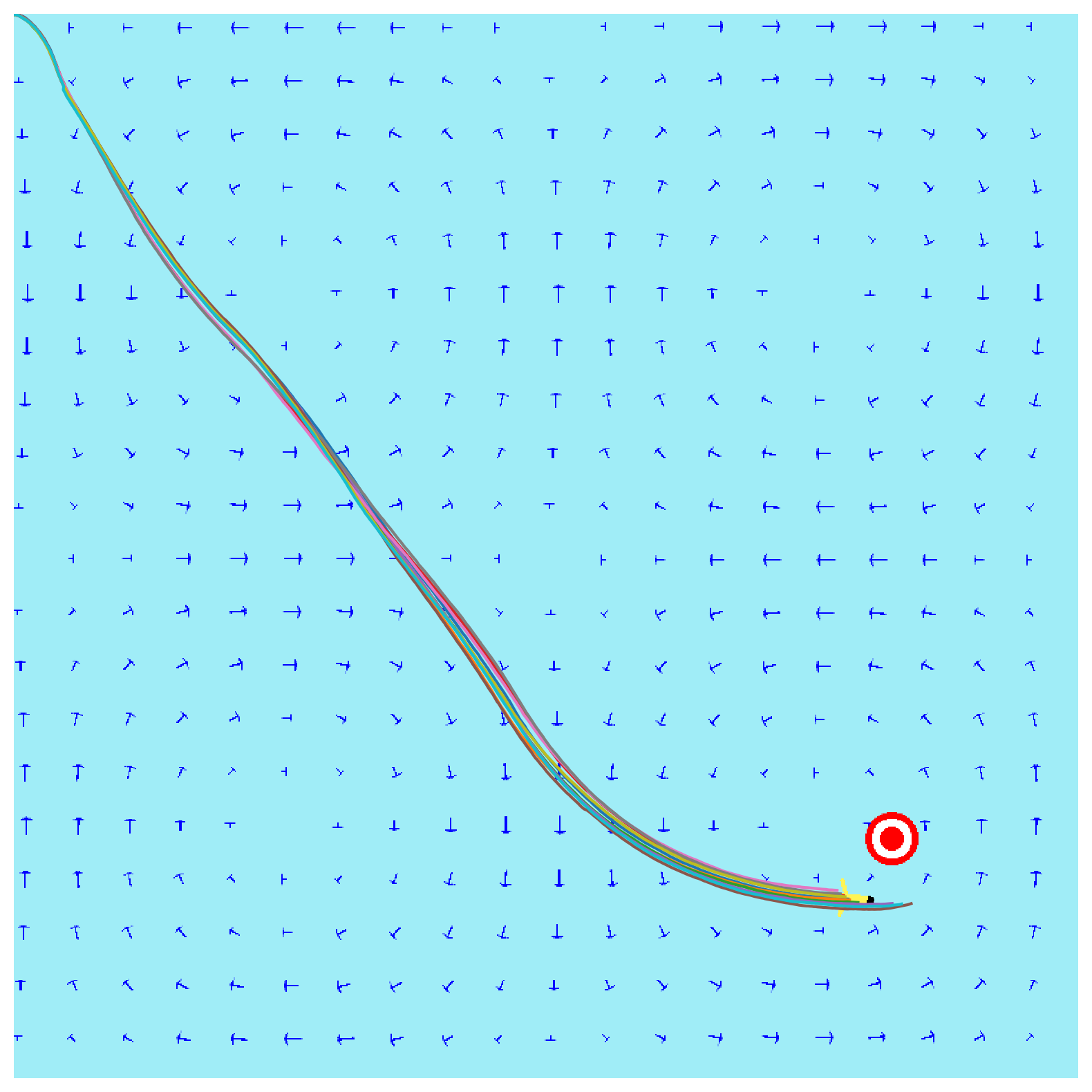}}
    \subfigure[Approximate policy iteration]
        {\label{fig:app_weak_traj}\includegraphics[width=0.325\textwidth]{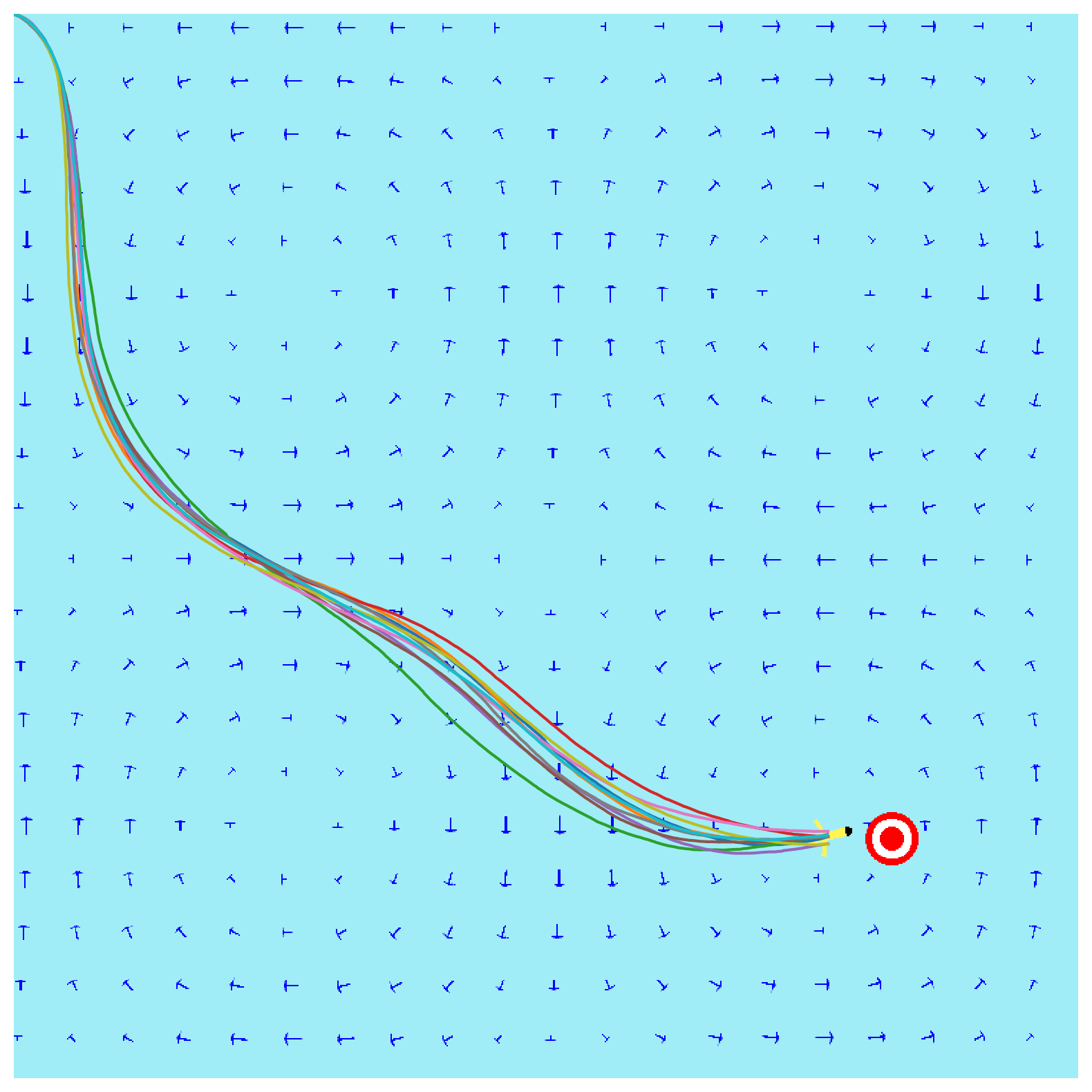}}
    \subfigure[Goal-oriented planner]
        {\label{fig:goal_weak_traj}\includegraphics[width=0.325\textwidth]{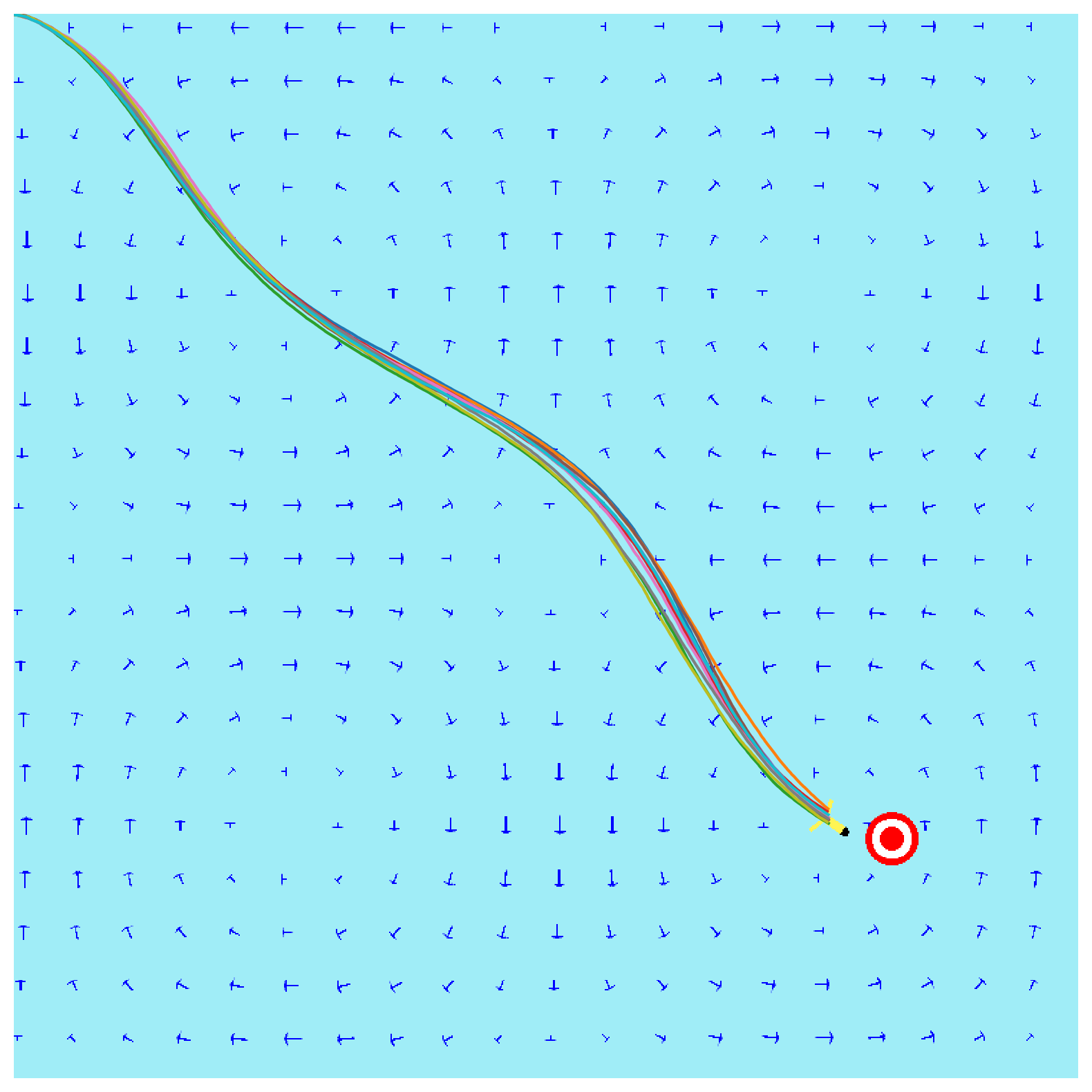}}\\
    \subfigure[Classic Policy Iteration]
        {\label{fig:exact_strong_traj}\includegraphics[width=0.325\textwidth]{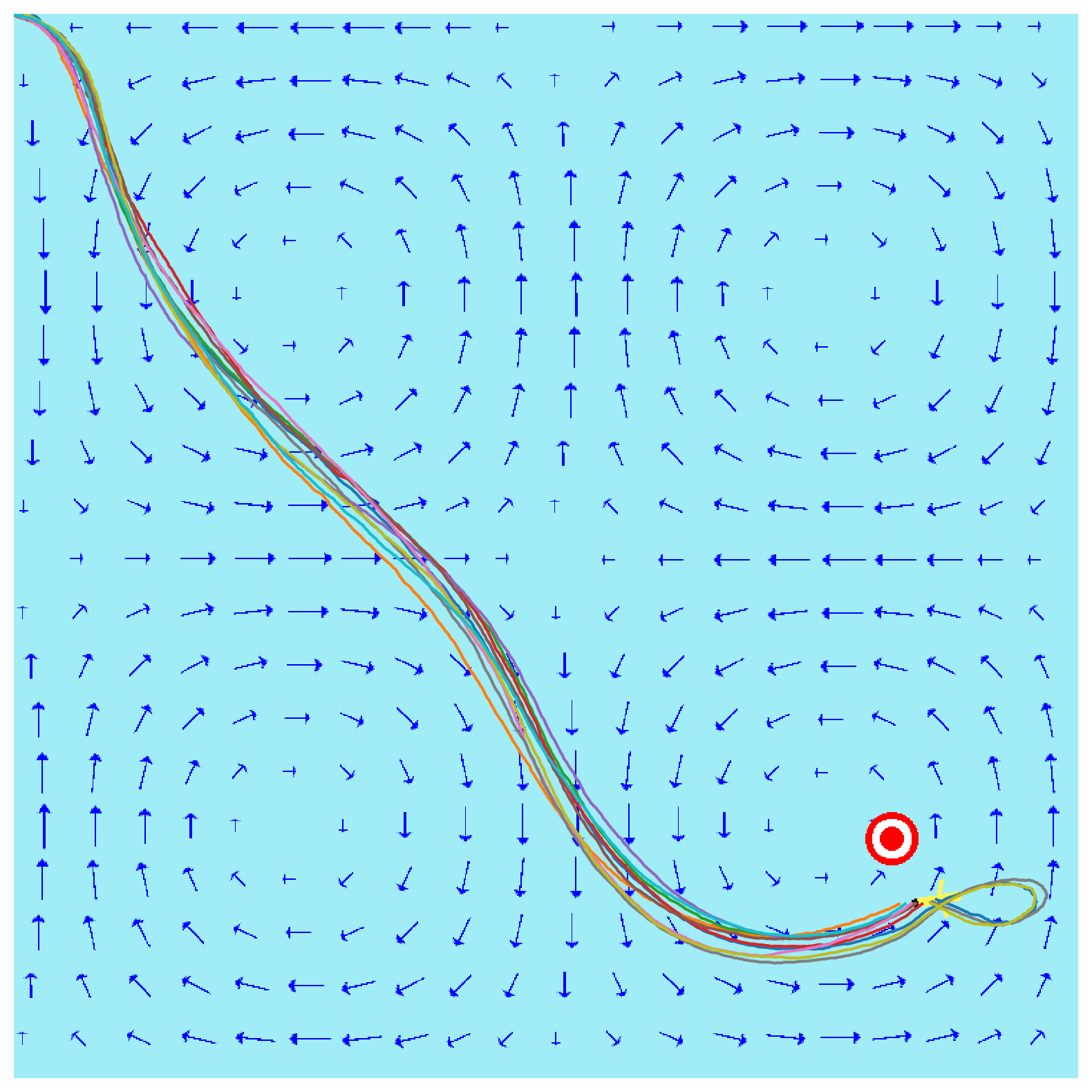}}
    \subfigure[Approximate policy iteration]
        {\label{fig:app_strong_traj}\includegraphics[width=0.325\textwidth]{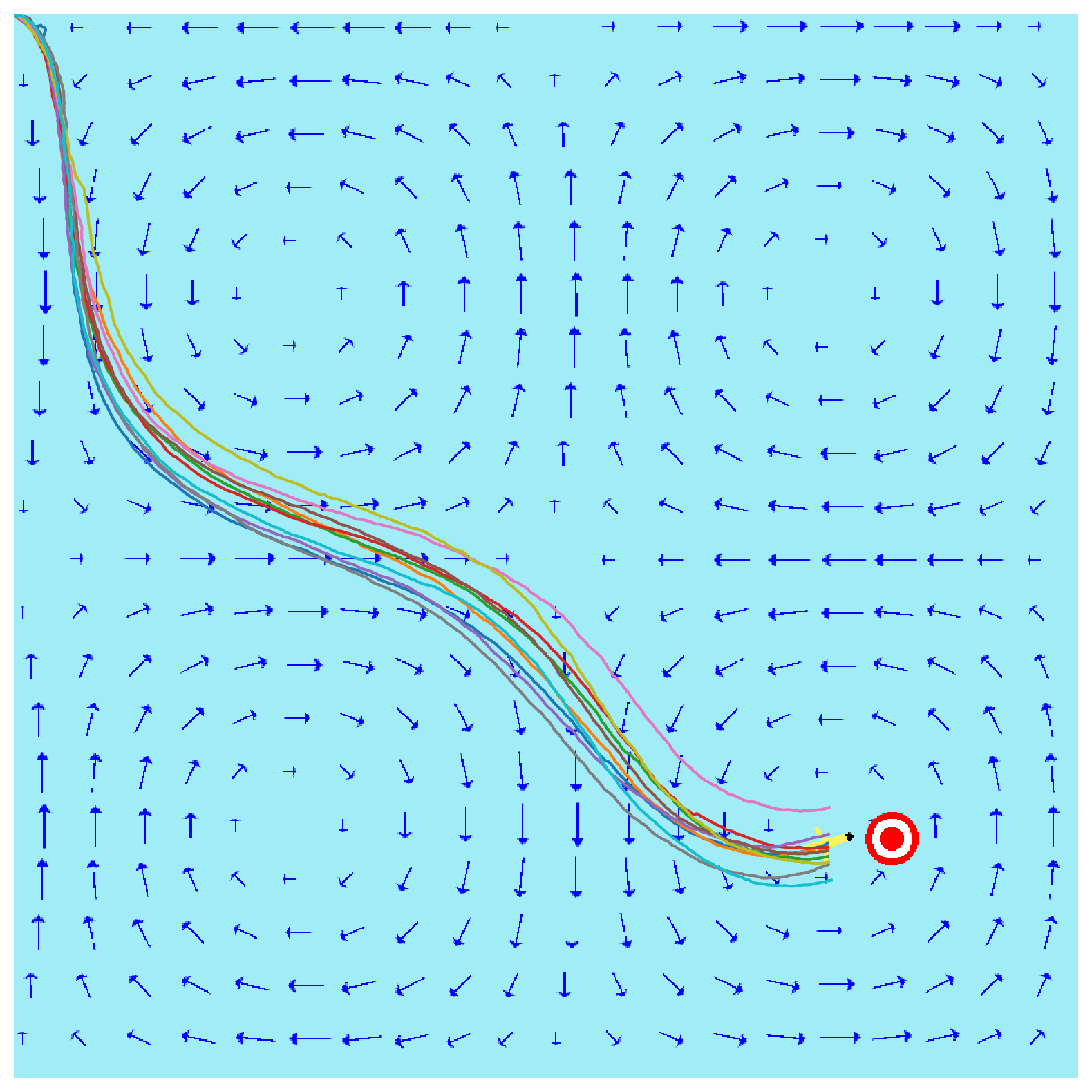}}
    \subfigure[Goal-oriented planner]
        {\label{fig:gaol_strong_traj}\includegraphics[width=0.325\textwidth]{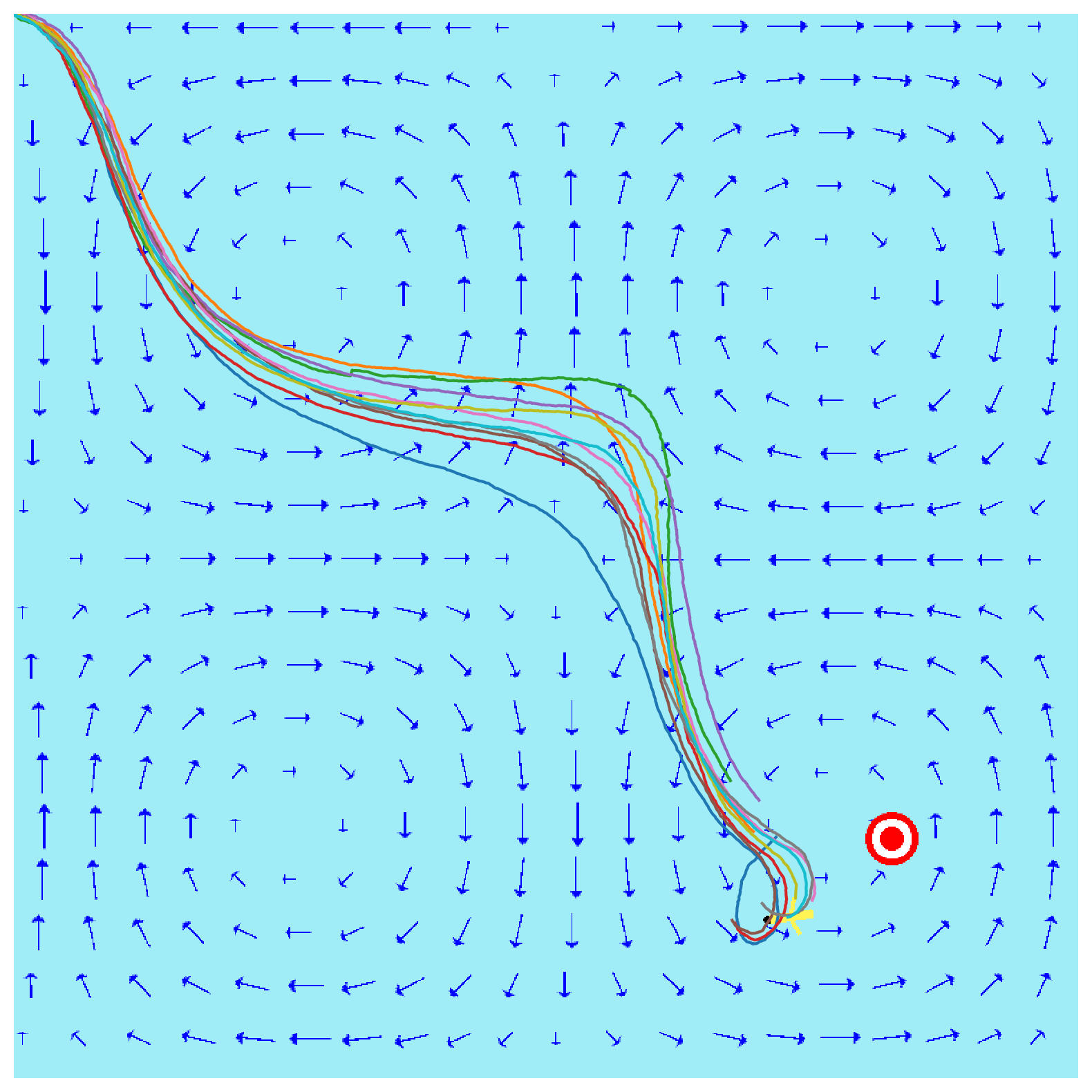}}\\
    \subfigure[Classic Policy Iteration]
        {\label{fig:exact_strong_obs_traj}\includegraphics[width=0.325\textwidth]{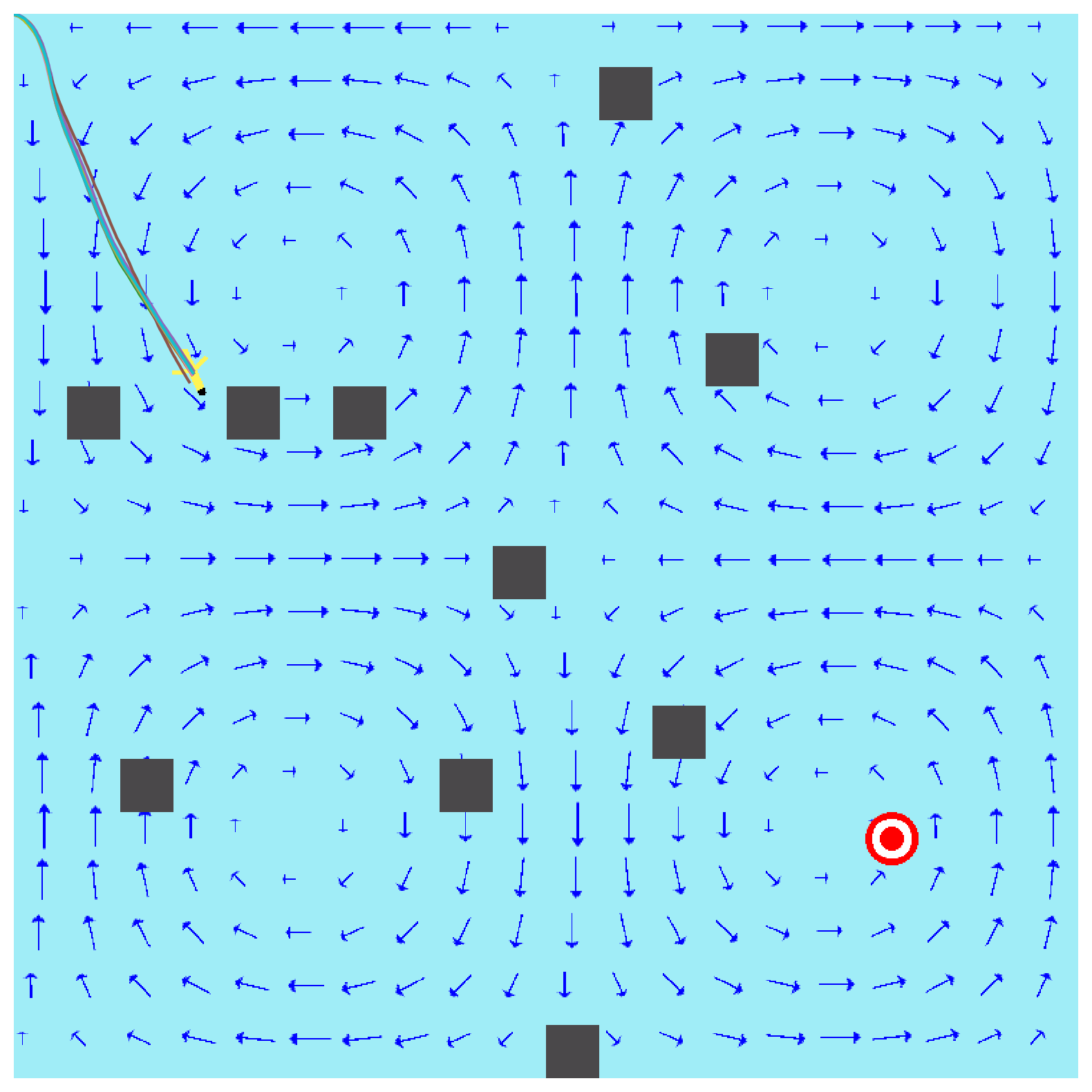}}
    \subfigure[Approximate policy iteration]
        {\label{fig:app_strong_obs_traj}\includegraphics[width=0.325\textwidth]{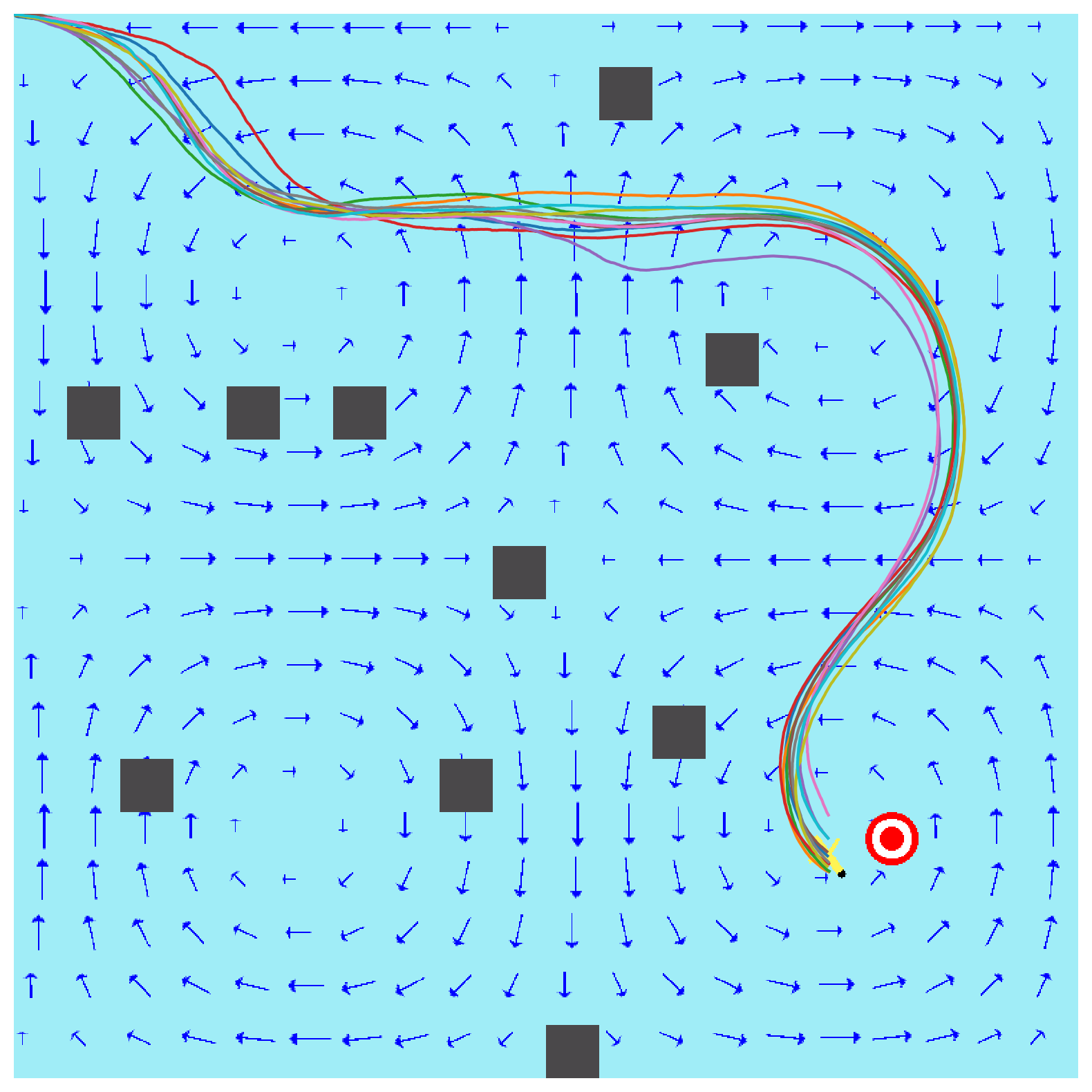}}
    \subfigure[Goal-oriented planner]
        {\label{fig:simple_strong_obs_traj}\includegraphics[width=0.325\textwidth]{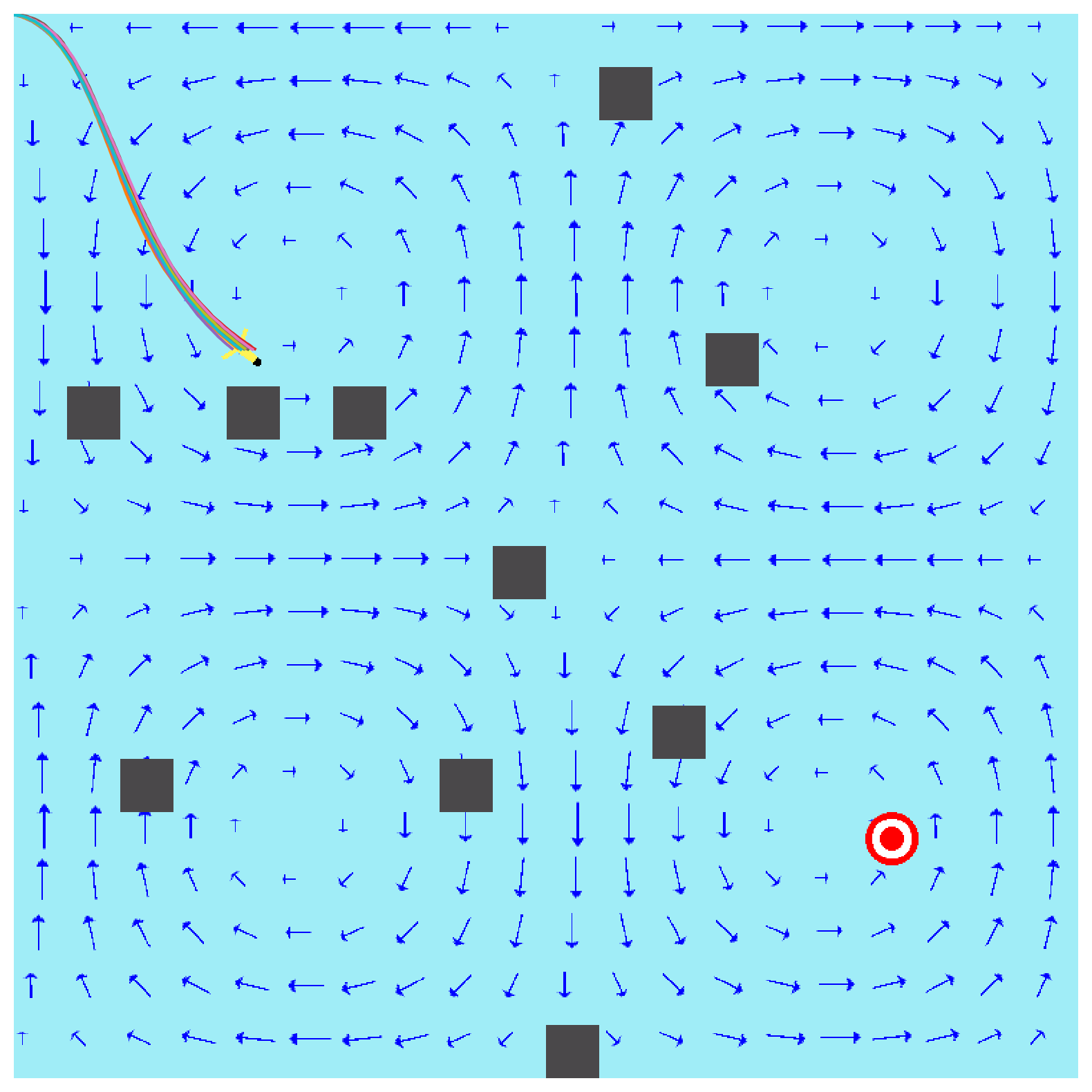}}
\caption{
\small{
Trajectory comparisons of $10$ trials under gyre disturbances. 
The top three figures demonstrate the paths under weak gyre disturbances, where $A=0.32$. Standard deviations of the disturbance velocity vector are set to $\sigma_x = \sigma_y = 1km/h$. The middle row shows the paths under stronger and more uncertain disturbances, where $A=0.8$ and $\sigma_x=\sigma_y=1.5km/h$. The last row demonstrates the paths with random obstacles (e.g., oil platforms or islets) in the environment.}
} 
\vspace{-8pt}
\label{fig:traj_gyre}
\end{figure}

In this section, we present experimental results for autonomous vehicle planning with an objective of minimizing time cost to a designated goal position under uncertain ocean disturbances. 
\RE{Although we focus on marine vehicles in the experiment, our method is general enough to other types of robots that suffer from environmental disturbances. 
For example, our approach can be directly applied to plan an energy-efficient path for aerial vehicles under the wind disturbances, where the MDP is a typical modeling method~\cite{al2013wind}.} 

\RE{
Our method is implemented using the \textit{FEniCS} package~\cite{AlnaesBlechta2015a}
a widely used FEM library.
Throughout the experiments, we use first-order Lagrange polynomials as the basis functions. 
}
We compare the proposed approach with two other baseline methods.
The first one is the classic MDP policy iteration (classic PI), which tessellates the continuous state space into grids, and the value function is represented by a look-up table~\cite{thrun2002probabilistic}.
This approach requires knowledge of the transition function of the MDP.
The second one is a goal-oriented planner which maintains the maximum vehicle speed and always keeps the vehicle heading angle towards the goal direction. 
The latter method has been widely used in navigating underwater vehicles due to its ``effective but lightweight" properties~\cite{ma2018data}.
We measure the performance in terms of the time cost of the trajectories.


\subsection{Evaluation with Gyre Model}
\subsubsection{Experimental Setup}
We first consider the task of navigating an autonomous underwater vehicle (AUV) to a designated goal area 
\Wording{subject to currents from a wind-driven ocean gyre.}
The gyre model is commonly used in analyzing flow patterns in large scale recirculation regions in the ocean \cite{blumberg1987description}. 
In this experiment, the dimension of the ocean surface is set to $20km \times 20km$. 
Similar to~\cite{kularatne2016time, kularatne2018optimal}, we use a velocity vector field $\mathbf{v}^d(s) = [v_{x}^{d}(s), v_{y}^{d}(s)]^{T}$ to represent the gyre disturbance at each location of the 2-D ocean surface. 
Its velocity components are given by $v_x^d(s) = -\pi A\sin{(\pi \frac{x}{e})}\cos{(\pi \frac{y}{e})}$ and $v_y^d(s) = \pi A\cos{(\pi \frac{x}{e})}\sin{(\pi \frac{y}{e})}$ respectively, where~$s = [x, y]^T$ is the location, $A$ denotes the strength of the current, and $e$ determines the size of the gyres. 

Due to estimation uncertainties~\cite{smith2013investigation}, the resulting $v_x^d$ and $v_y^d$ 
\Wording{do not accurately}
reflect the actual dynamics of the ocean disturbance.
For effective and accurate planning, 
these uncertainties need to be considered,
\Wording{and they are modeled as additive environmental noises.}
To reduce the complexities on modeling and computing, we adopt the existing approximation methods~\cite{kularatne2018optimal, huynh2015predictive, liu2018solution} and assume the noise along two dimensions $v_x^d$ and $v_y^d$ follows independent Gaussian distributions \Wording{$\Tilde{\mathbf{v}}^d(s) = \mathbf{v}^d(s) + \mathbf{w}(s)$},
where $\Tilde{\mathbf{v}}^d(s)$ denotes the velocity vector at position $s$ after introducing the uncertainty and $\mathbf{w}(s)$ is the noise.
The components of the noise vector are given by 
\begin{equation}\label{eq:uncertainty}
w_x(s) \sim \mathcal{N}(0, \sigma^2_x(s)),
w_y(s) \sim \mathcal{N}(0, \sigma^2_y(s)),
\end{equation}
where $\mathcal{N}(\cdot, \cdot)$ denotes Gaussian distribution, and $\sigma^2_x$ and $\sigma^2_y$ are the noise variance for each component, respectively.

The state is defined as the position of the AUV, i.e., $s \in \mathbb{S} \subseteq \mathbb{R}^2$.
The actions are defined by the vehicle moving at its maximum speed (determined by the vehicle's capability) $v_{max} = 3km/h$ towards $Q=8$ desired heading directions in the fixed world frame~$\mathbb{A}= \{a_i | i \in \{1 \dots Q\}\}$, where~$a_i = [v_{max}\cos(\frac{2\pi i}{Q}), v_{max}\sin(\frac{2\pi i}{Q})]^T$ describes the velocity vector of the vehicle.
The vehicle's motion is affected by both the vehicle's action and the uncertain external disturbance.
Thus, the next state~$s'$ of the vehicle starting at state~$s$ after following the desired action~$a$ for a fixed time interval~$dt$ is given by~$s' = (a + \Tilde{\mathbf{v}}^d(s))dt$.
Since the velocity vector of the ocean current is perturbed by the additive Gaussian noise, 
the next state is a random variable, and the corresponding transition probability is given by 
\begin{equation}\label{eq:transition-function}
    \mathcal{T}(s, a; s') = \mathcal{N}(\mu(s), \mbox{diag}[\sigma_x^2dt^2, \sigma_y^2dt^2]),
\end{equation}
where $\mu(s) = s + (a + \mathbf{v}^d(s))dt$.
We have assumed that the ocean disturbances are constant near the current state $s$, and executing the action will not carry the vehicle too far away from the current state. 
To satisfy this assumption, we set the action execution time to a relatively small duration~$dt = 0.1h$.
The reward is one when the current state is the goal state, and zero otherwise.
Because we are interested in minimizing the travel time, we set the reward discount factor as $\gamma = 0.9$.
\RE{Also, we set goal and obstacle areas to be $1km \times 1km$ regions, and the states within these areas are absorbing states, i.e., the vehicle cannot transit to any other states if the current state is within these areas.} 
Thus, the boundary conditions 
within the goal and obstacle areas have values of $\frac{1}{1-\gamma} = 10$ and $\frac{0}{1-\gamma}=0$, respectively.

To model the transition function of the classic MDP planner, we follow the approach commonly used in AUV planning literature~\cite{hollinger2016learning}. 
Specifically, each state $s$ is represented by a regular grid, and the next state transition probabilities are only assigned to its 8-connected neighbors based on Eq.~\eqref{eq:transition-function}.

\subsubsection{Trajectories and Time Costs}
\begin{figure}[t] 
    \centering
    \subfigure[]
        {\label{fig:time_cost}\includegraphics[width=0.476\textwidth]{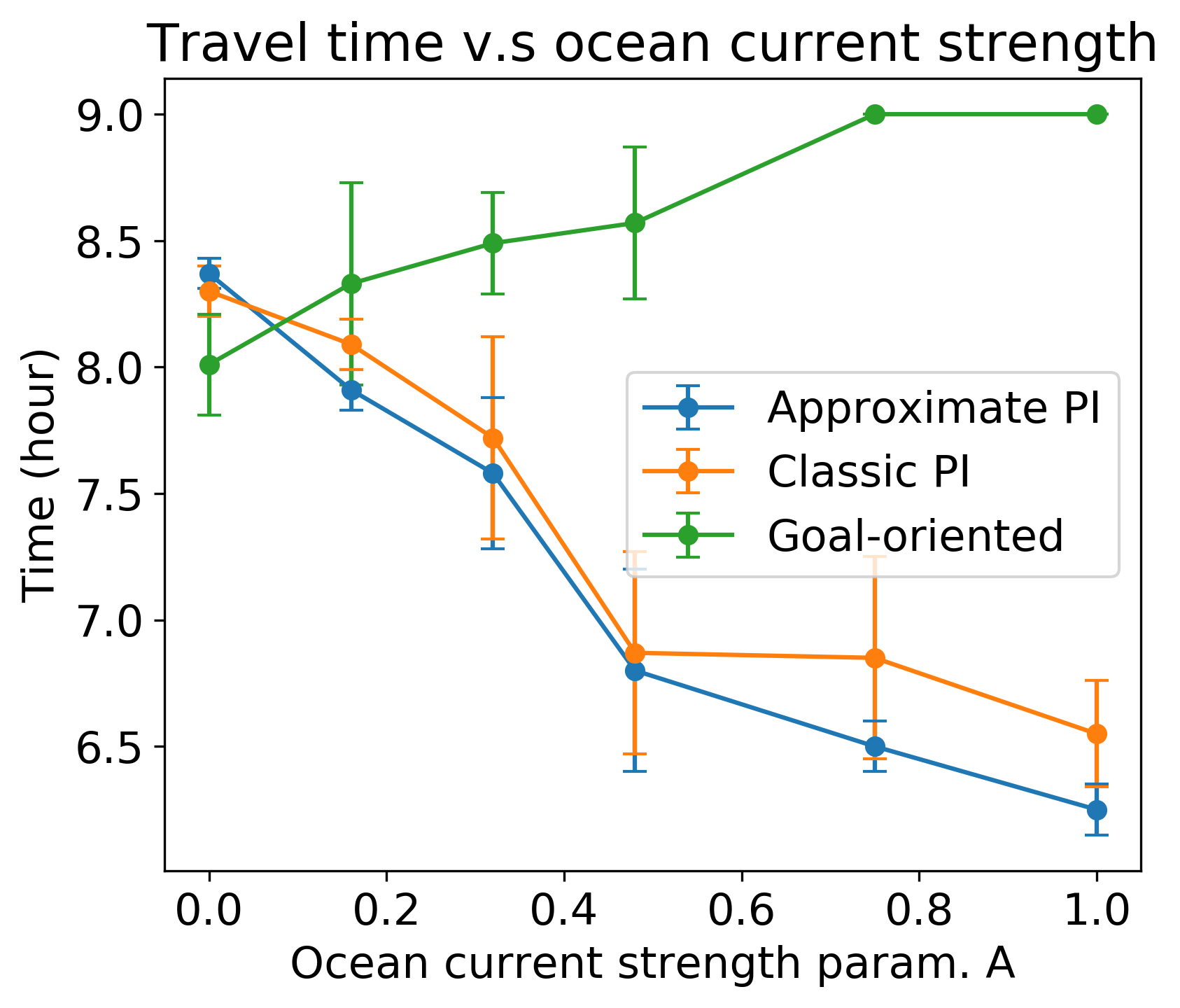}}
    \vspace{-10pt}
    \subfigure[]
        {\label{fig:traj_cost}\includegraphics[width=0.49\textwidth]{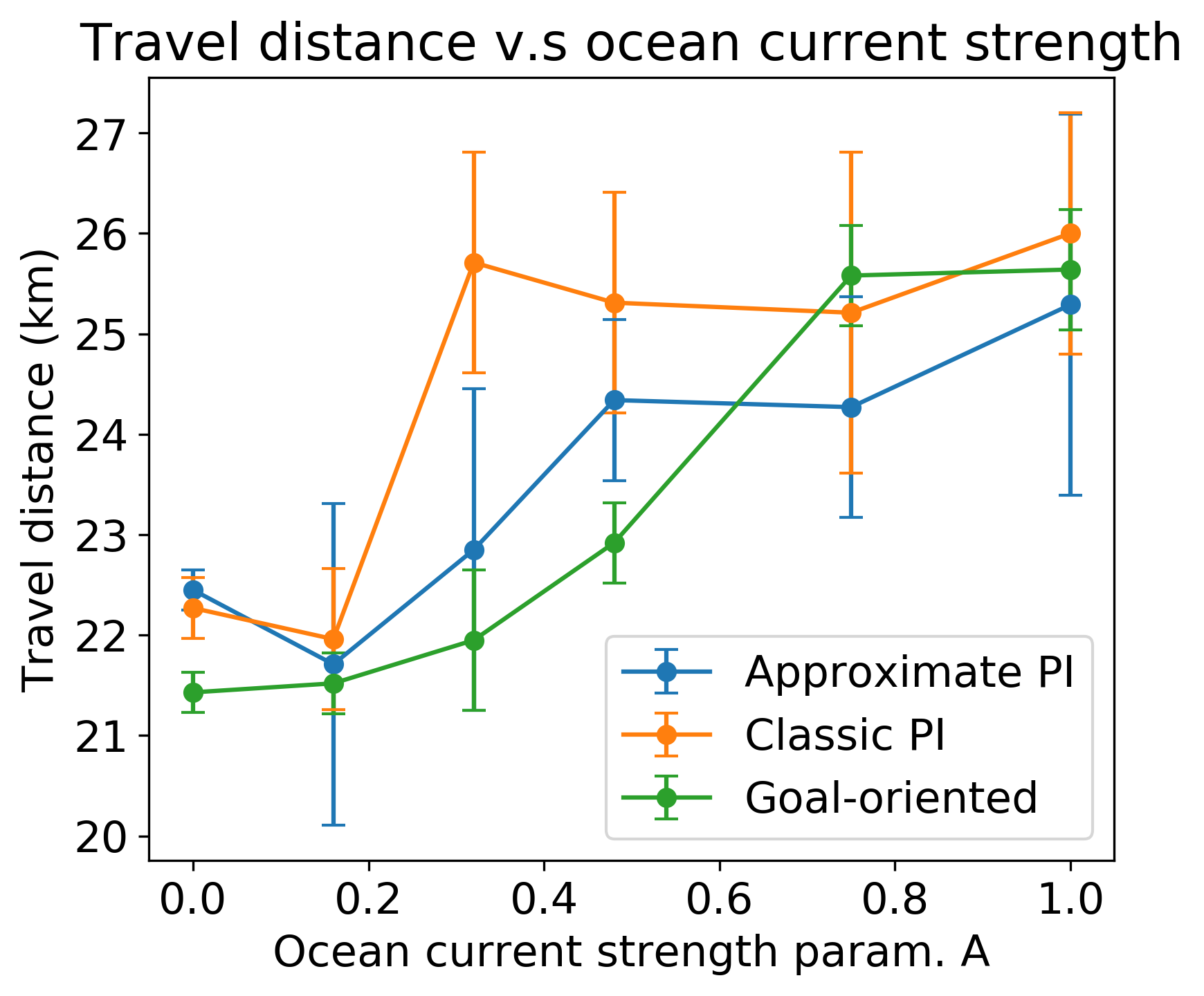}}
\caption{\small{The time costs and trajectory lengths \Wording{yielded by} the three methods \Wording{with} different disturbance strengths. These results are averaged over 10 trials.}}
\vspace{-5pt}
\label{fig:performance_gyre} 
\end{figure}

\begin{figure}[t]
    \centering
    \subfigure[]
        {\label{fig:exact_weak_value}\includegraphics[width=0.47\textwidth]{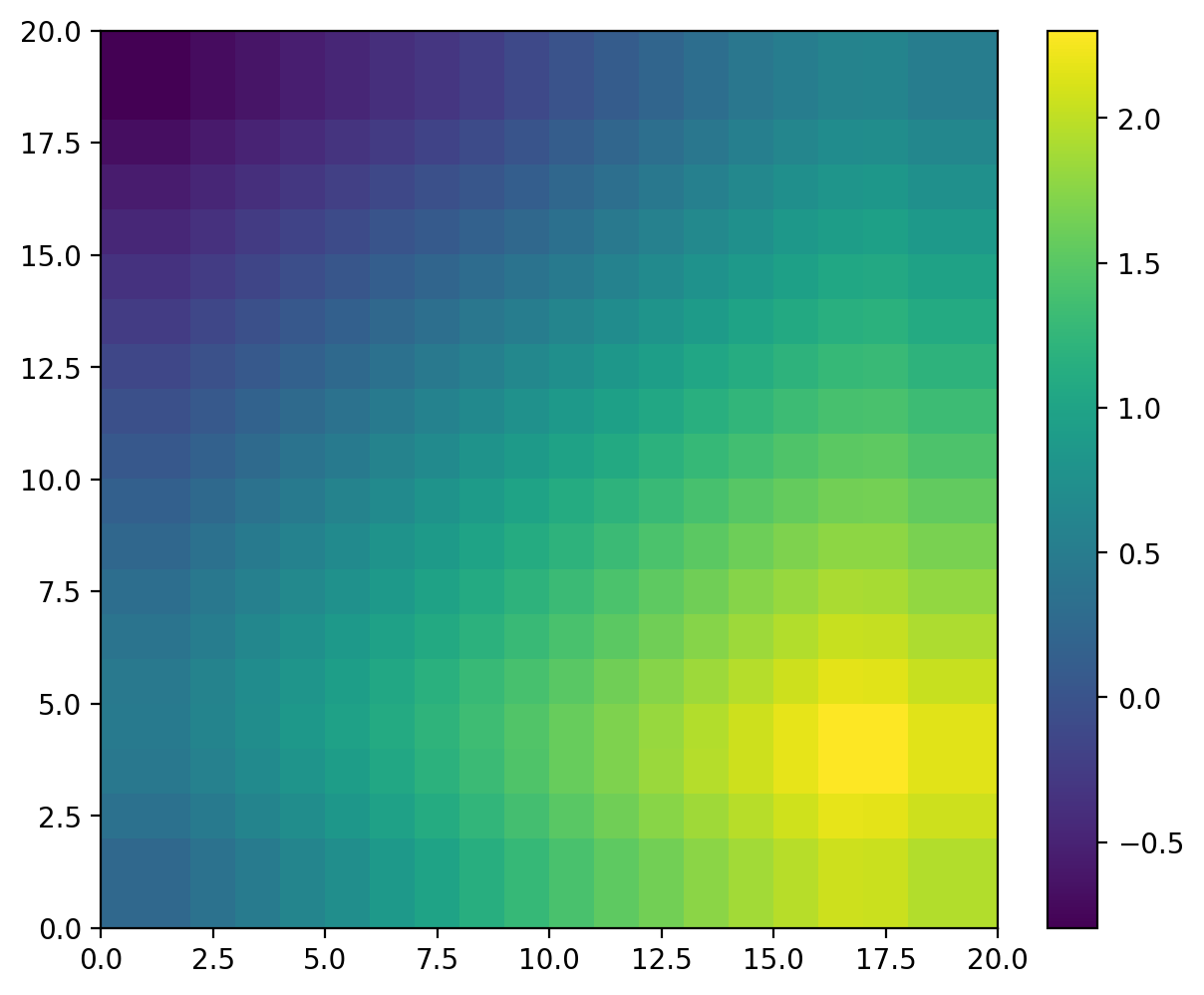}
        }
    \vspace{-5pt}
    \subfigure[]
        {\label{fig:app_weak_value}\includegraphics[width=0.45\textwidth]{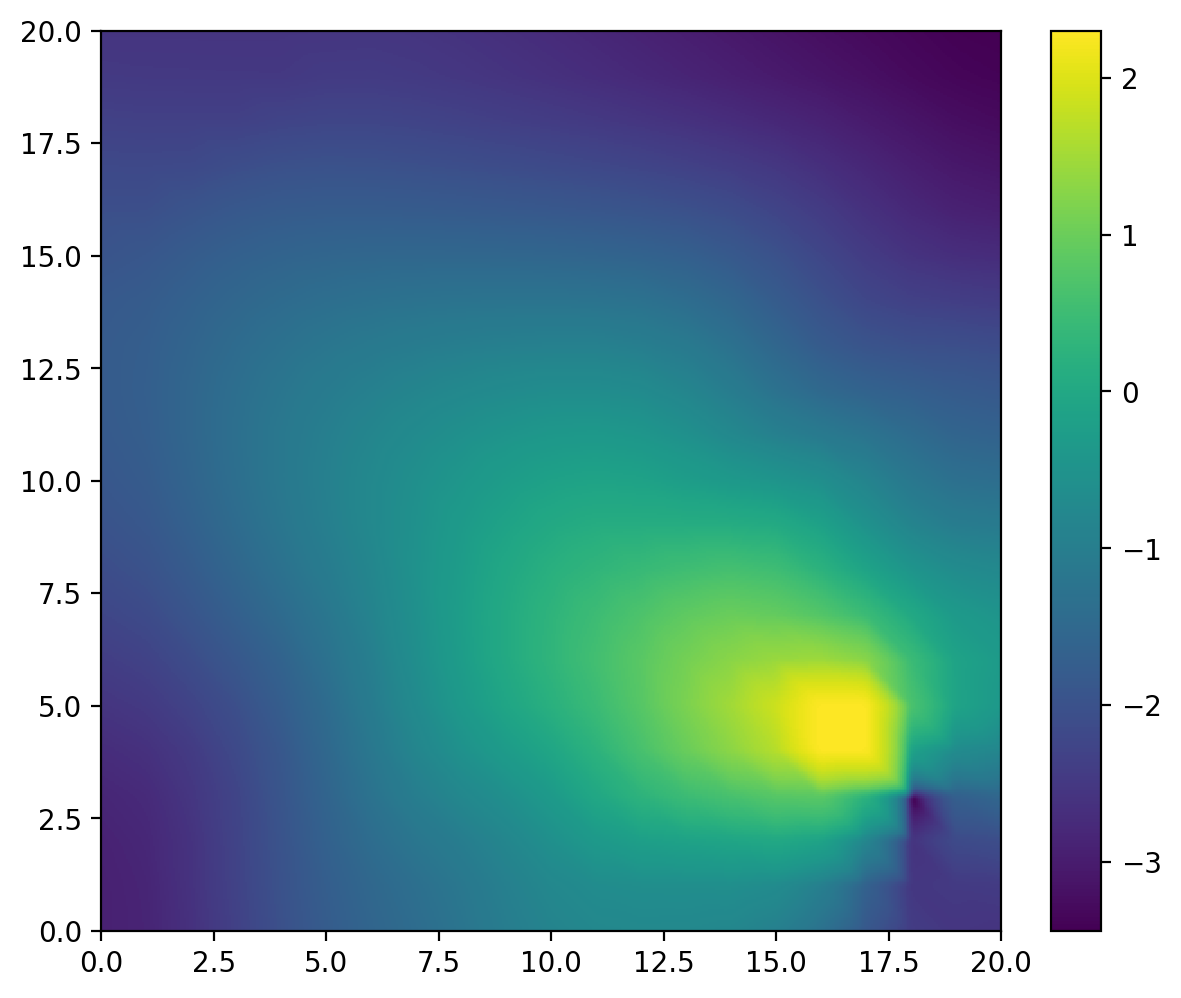}}\\
    \vspace{-5pt}
    \centering
    \subfigure[]
        {\label{fig:exact_strong_value}\includegraphics[width=0.47\textwidth]{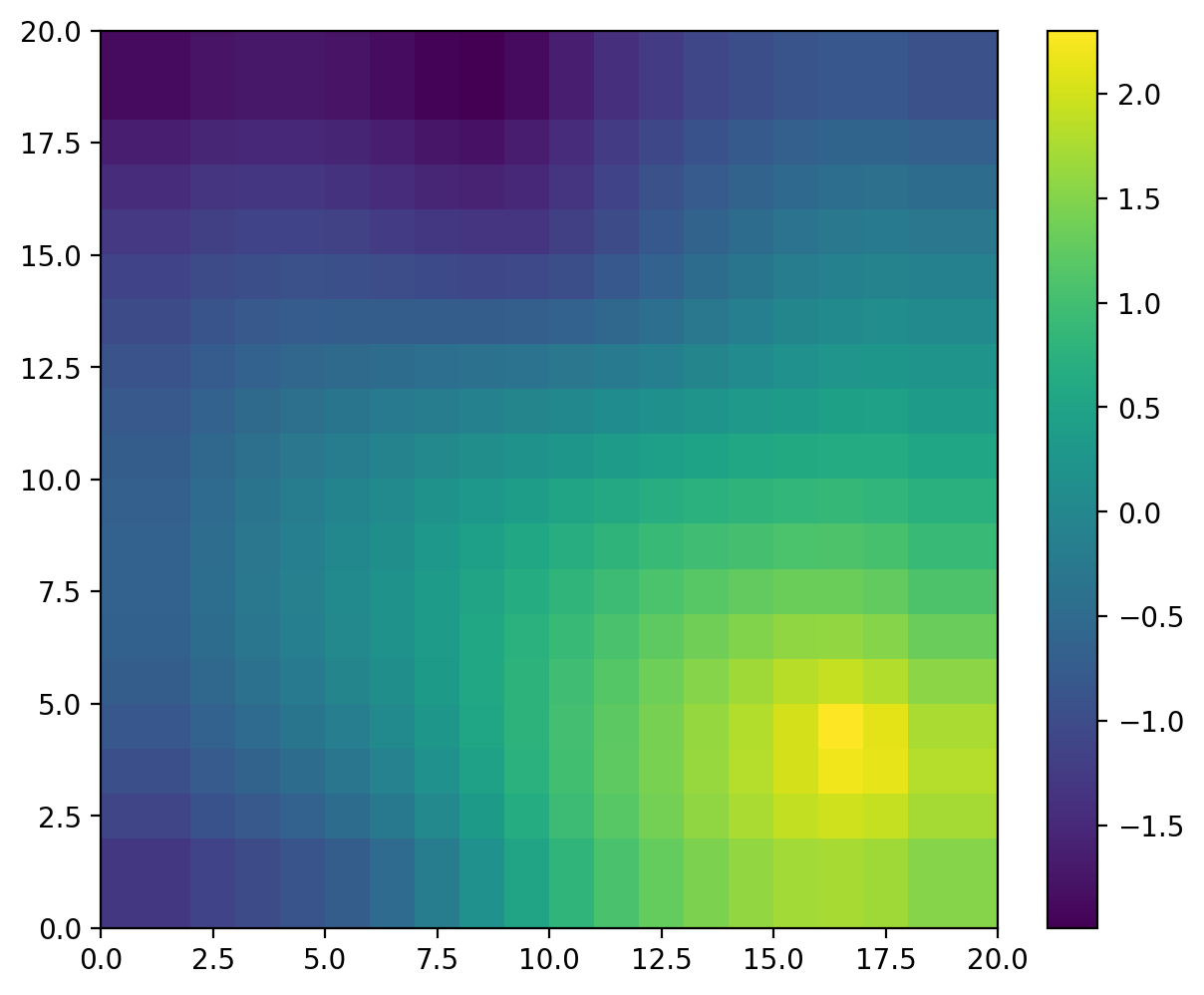}}
    \vspace{-5pt}
    \centering
    \subfigure[]
        {\label{fig:app_strong_value}\includegraphics[width=0.47\textwidth]{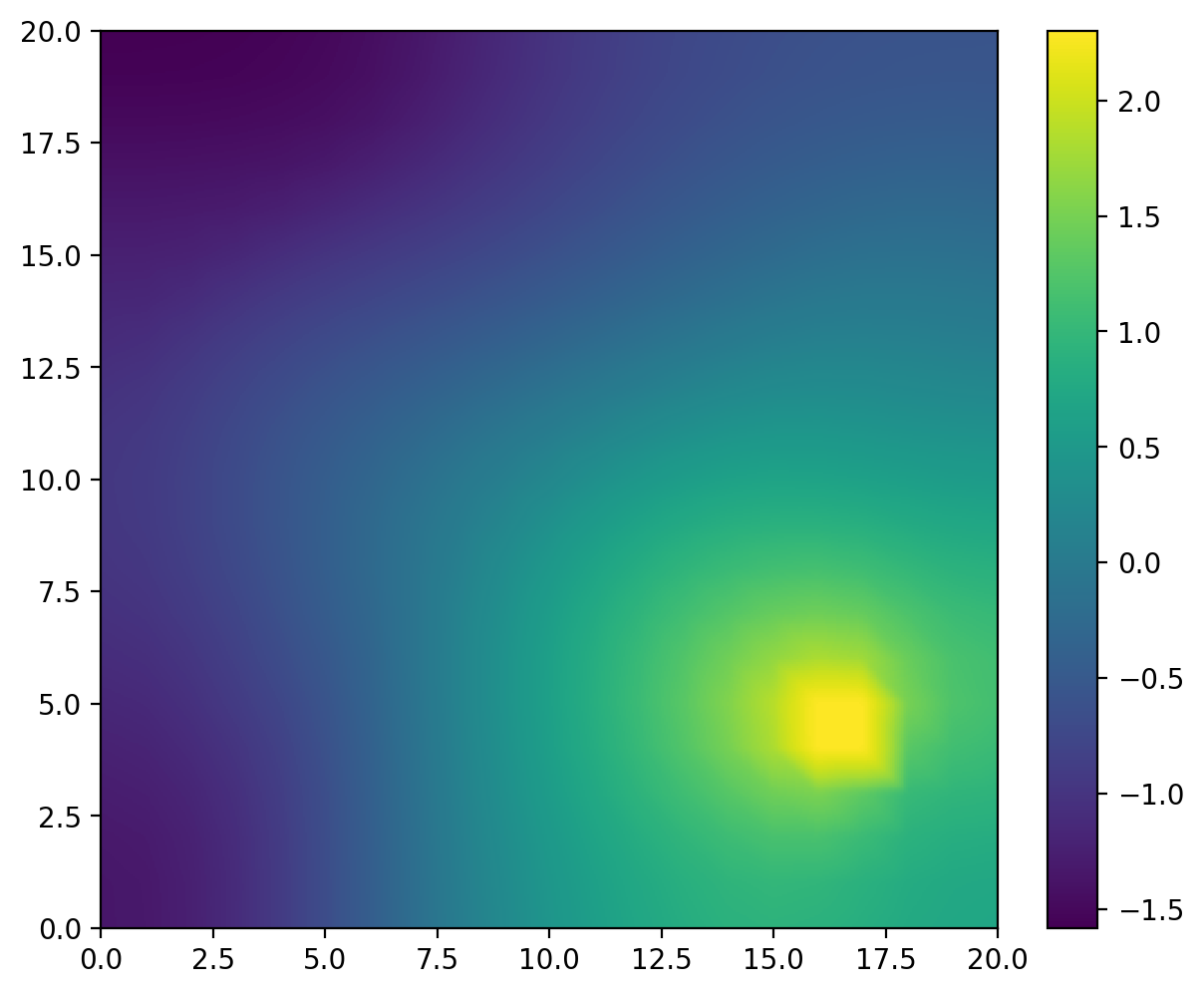}}\\
    \vspace{-5pt}
    \centering
    \subfigure[]
        {\label{fig:exact_strong_obs_value}\includegraphics[width=0.47\textwidth]{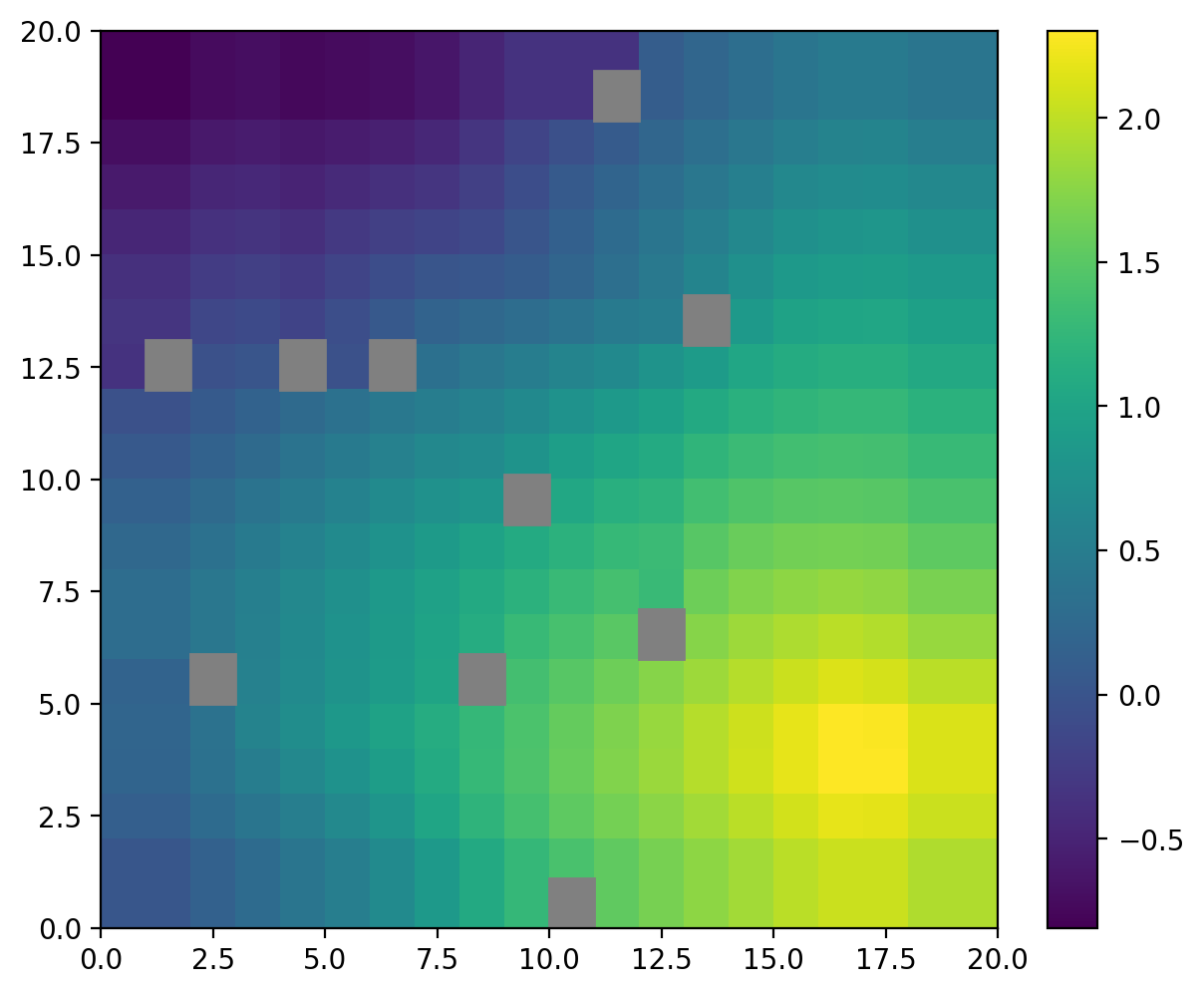}}
    \vspace{-5pt}
    \centering
    \subfigure[]
        {\label{fig:app_strong_obs_value}\includegraphics[width=0.46\textwidth]{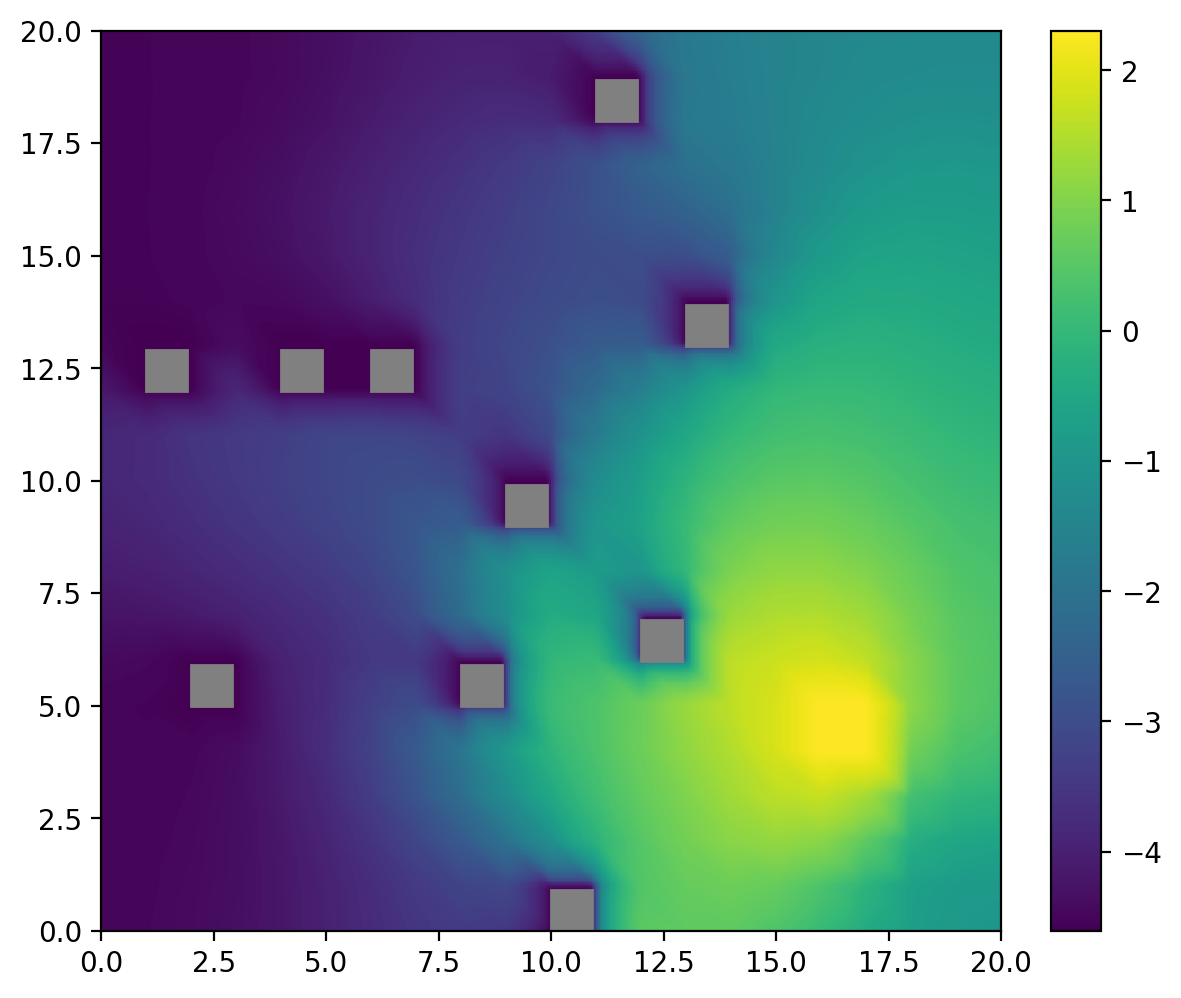}}\\
\caption{\RE{\small{Value functions (values are in log scale) calculated from 
PI on the left and approximate policy iteration on the right.
 The environment parameters and settings are the same as for 
 Fig.~\ref{fig:traj_gyre}.}}
}
\vspace{-5pt}
\label{fig:value_funct} 
\end{figure}

\begin{table}[t]
\centering
\resizebox{0.98\columnwidth}{!}{%
    \begin{tabular}{lcccc}
          $\bf{A=0.0}$& PI (400 states) & $h=0.5km$ & $h=1km$ & $h=2km$\\
         \hline
          Time cost & $8.3 \pm 0.1$ & $\bf{8.23 \pm 0.16}$ & $8.37 \pm 0.3$ & $8.36 \pm 0.1$\\
          Traj. len. & $22.27 \pm 0.3$ & $\bf{22.03 \pm 0.5}$ & $22.45 \pm 0.2 $ & $ 22.46 \pm 0.2$\\
          \addlinespace[0.2cm]
          $\bf{A=0.16}$ &&&\\
          \hline
          Time cost & $8.09 \pm 0.1$ & $\bf{7.9 \pm 0.16}$ &$7.91 \pm 0.08$ & $8.09 \pm 0.12$\\
          Traj. len. & $21.96 \pm 0.7$ & $21.8 \pm 0.14$ & $\bf{21.71 \pm 1.6}$ & $23.58 \pm 0.5$\\
          \addlinespace[0.2cm]
          $\bf{A=0.32}$ &&&\\
          \hline
          Time cost & $7.72 \pm 0.4$ & $\bf{7.52 \pm 0.11}$ & $7.58 \pm 0.3$ & $7.64 \pm 0.29$\\
          Traj. len. & $25.71 \pm 1.1$ & $\bf{22.34 \pm 0.86}$ & $22.85 \pm 1.6$ & $23.85 \pm 1.1$\\
          \addlinespace[0.2cm]
          $\bf{A=0.48}$ &&&\\
          \hline
          Time costs & $6.87 \pm 0.4$ & $\bf{6.51 \pm 0.1}$ & $6.8 \pm 0.4$ & $ 7.6 \pm 0.1 $\\
          Traj. len. & $25.31 \pm 1.1$ & $\bf{23.92 \pm 1.1}$ & $24.34 \pm 0.8$ & $ 24.43 \pm 1.1 $\\
          \addlinespace[0.2cm]
          $\bf{A=0.75}$ &&&\\
          \hline
          Time costs & $6.85 \pm 0.4$ & $6.6 \pm 0.5 $& $\bf{6.5 \pm 0.4}$ & $6.82\pm0.12$\\
          Traj. len. & $25.21 \pm 1.6$ & $24.5 \pm 0.86$ & $\bf{24.34 \pm 1.1}$ & $24.53 \pm 1.8$\\
          \addlinespace[0.2cm]
          $\bf{A=1.0}$ &&&\\
          \hline
          Time costs & $6.55 \pm 0.21$ & $\bf{6.08 \pm 0.12}$ &$ 6.25 \pm 0.1$ & $ 6.51 \pm 0.16$ \\
          Traj. len. & $26.00 \pm 1.2$ & $ 25.74 \pm 1.03$&$ 25.29 \pm 1.0$ & $\bf{24.00\pm 1.8}$ \\
    \end{tabular}%
}
    \caption{\small{Time and trajectory costs averaged over $10$ trials with different ocean current strengths $A$.
    \RE{The statistics of classic policy iteration (PI) and our method with different resolution parameters $h$ are shown in each column.}
    The best performing statistics are highlighted with a bold font.}}
    \label{tab:discretization}
\end{table}

\begin{figure*}[t] 
\centering
    \subfigure[]
        {\label{fig:real_exact}\includegraphics[width=0.25\textwidth]{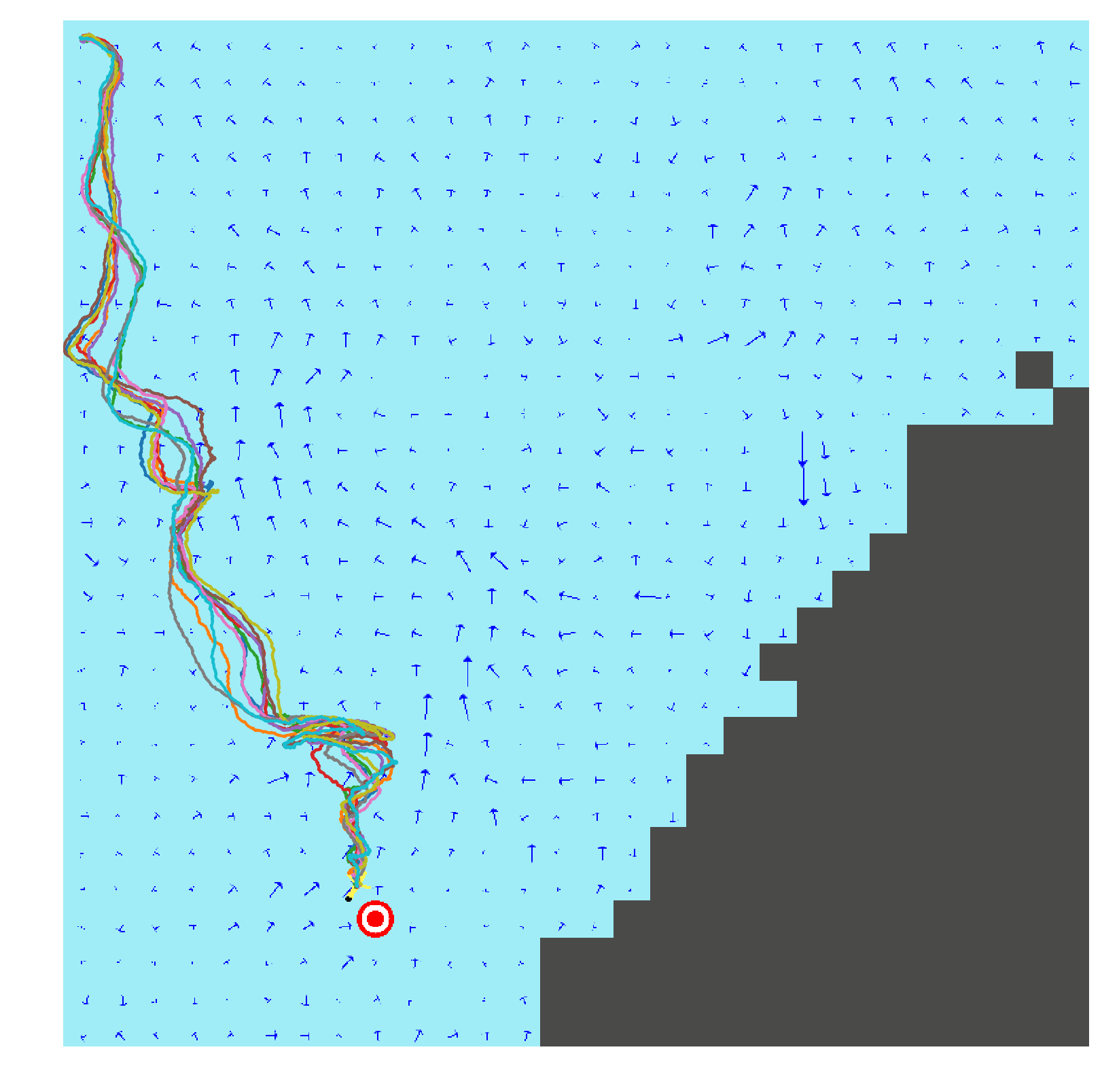}}
        \quad
    \subfigure[]
        {\label{fig:real_app}\includegraphics[width=0.26\textwidth]{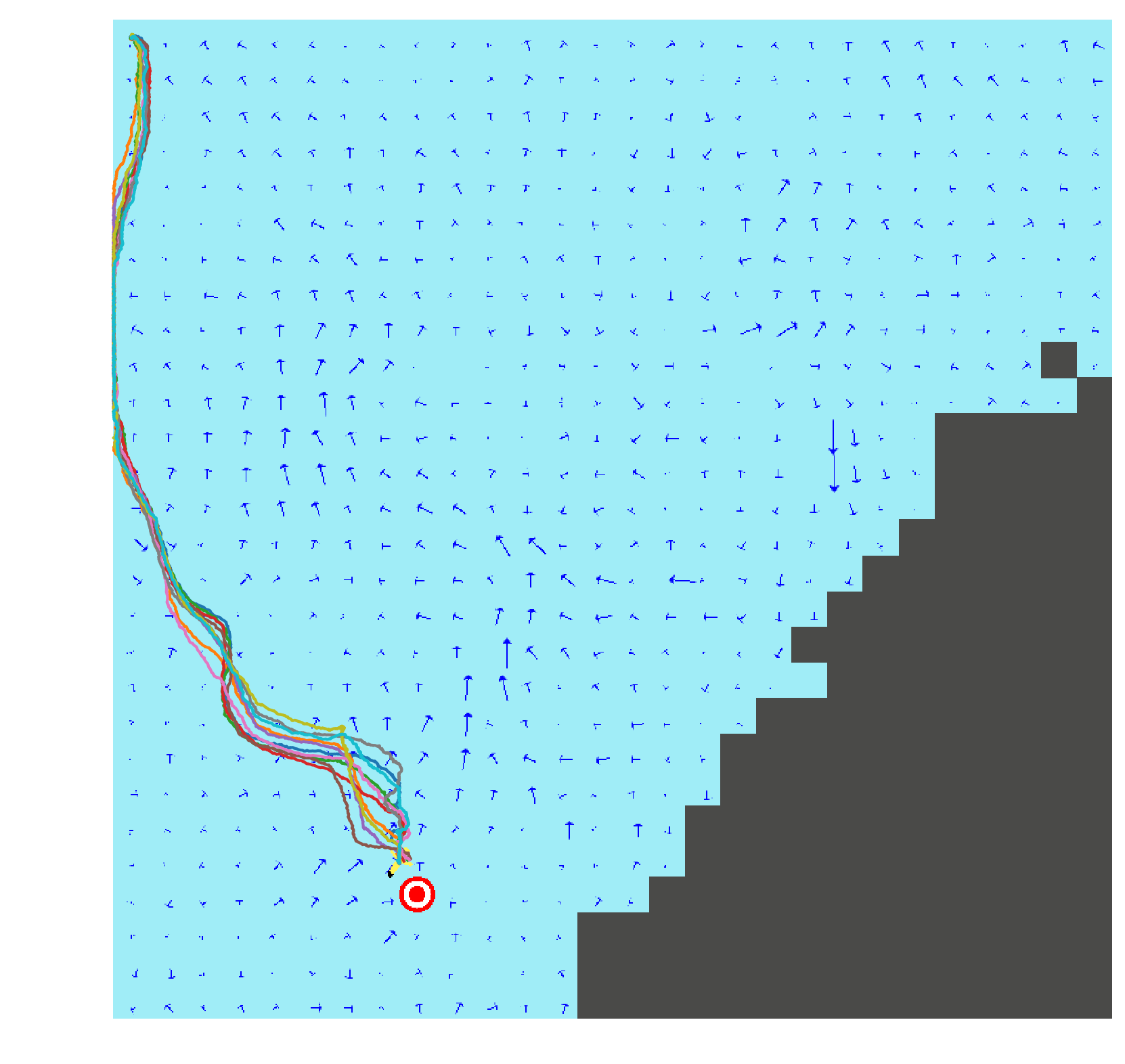}}
        \quad
    \subfigure[]
        {\label{fig:exact_strong_value}\includegraphics[width=0.25\textwidth]{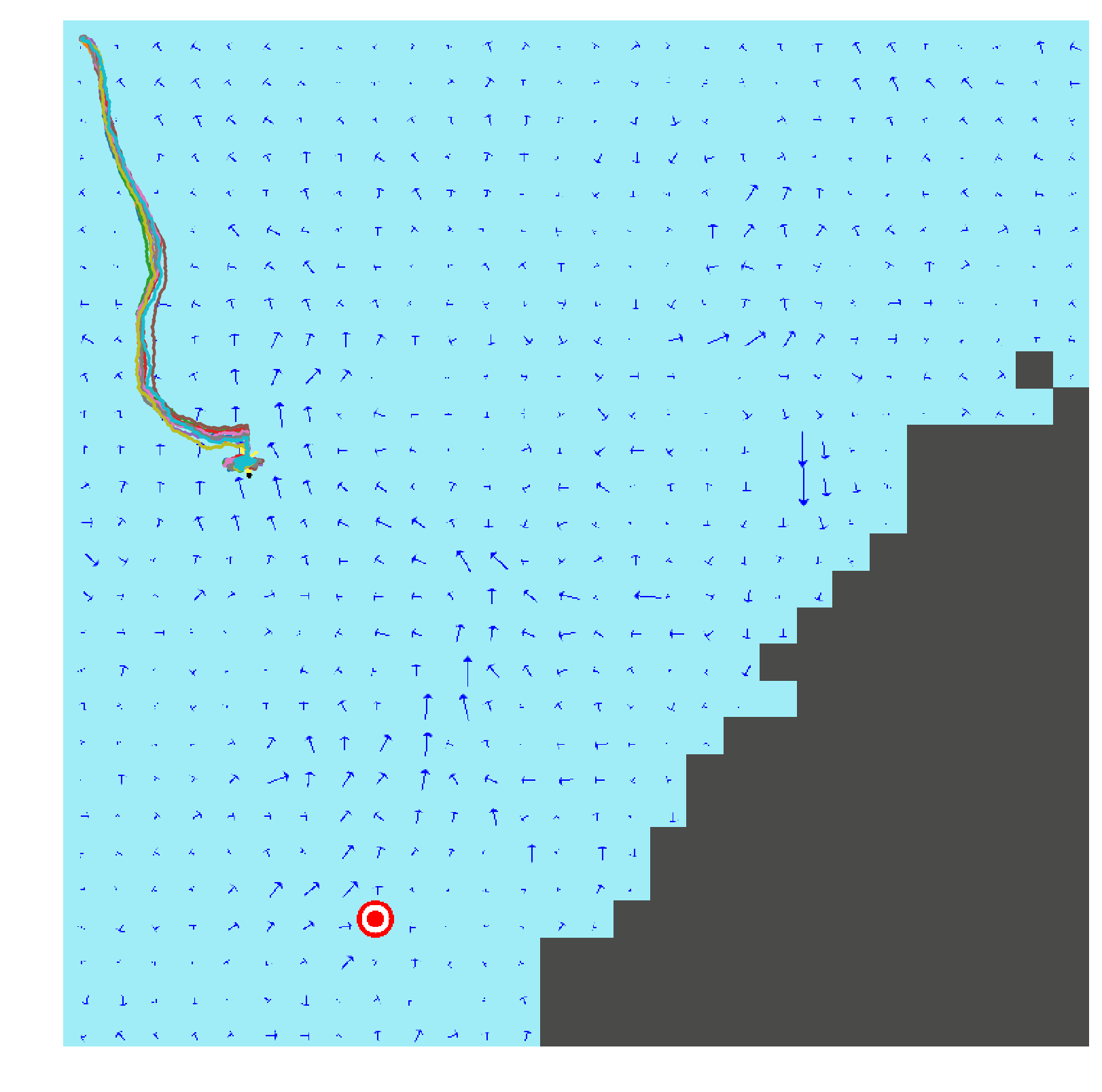}}
        \vspace{-5pt}
        \caption{\small{Trajectories from the three methods using ocean data. (a) Classic policy iteration; (b) Approximate policy iteration; (c) Goal-oriented planner. }}
        \label{fig:roms_traj} 
        \vspace{-8pt}
\end{figure*}

We present the performance evaluations in terms of trajectories and time costs. 
For the classic MDP planner (classic PI), we use $20 \times 20$ grids, and each cell has a dimension of $1km \times 1km$.  
To make a fair comparison, we use $20 \times 20$ state points for our method (approximate PI), and the corresponding resolution parameter is $h=1km$.
We set $s=10$ which generates four gyre areas. 
To simulate the uncertainty of estimations during each experimental trial, we sample Gaussian noises from Eq.~\eqref{eq:uncertainty} and add them to the calculated velocity components of the gyre model. 

We examine the trajectories of different methods under weak and strong disturbances in Fig.~\ref{fig:traj_gyre}. 
We run each method for 10 times. The colored curves represent accumulated trajectories of multiple trials. 
For the weak disturbance, parameter $A$ is set to $0.32$ and the \Wording{standard deviation of the noise} is set to $1km/h$ . 
The resulting maximum ocean current velocity is around $1km/h $, which is smaller than the vehicle's maximum velocity of $3km/h$. 
In this case, the vehicle has the capability of going forward against ocean currents. 
In contrast, the strong ocean current \Wording{has a} maximum velocity of $2.5km/h$, which is similar to the maximum vehicle velocity. 
By comparing the first and second rows of Fig.~\ref{fig:traj_gyre}, we can easily observe that under  weak ocean disturbances, the three methods produce trajectories that \Wording{converge on the target and look similar.}
This is because the weak disturbance results in smaller vehicle motion uncertainty. 
The goal-oriented planner has the shortest travel distance, but its travel time is the highest (Fig.~\ref{fig:goal_weak_traj} and Fig.~\ref{fig:performance_gyre}) under the weak disturbance.
On the other hand, the MDP policies can leverage the fact that moving in the same direction as the ocean current allows the vehicle to obtain a faster net speed and thus results in a shorter time to the goal.
Specifically, with strong disturbances, instead of taking a more direct path towards the goal position, the MDP policies take a longer path but with a faster speed by following the direction of the ocean currents.
In contrast, the goal-oriented planner is not able to reach the goal subject to the strong disturbance within the time budget (9h). 
Planning \Wording{subject to} the appearance of obstacles is shown in the last row of Fig.~\ref{fig:traj_gyre}. 
We can observe that, among the three methods, only ours successfully navigates to the goal area \Wording{with respect to} 
the strong ocean currents. 
Our method attempts to minimize the travel time \Wording{while} avoiding possible collision.
It first travels against the current to find a safe region and then follows the ocean current to advance to the goal area.
 
We further quantitatively evaluate the averaged time costs and trajectory lengths of the three methods \Wording{subject to} different disturbance strengths with fixed $\sigma_{x}=\sigma_{y}=1km/h$,  as illustrated in Fig.~\ref{fig:performance_gyre}. 
The overall performance for the proposed approximate policy iteration is consistently better than the other two methods in terms of the time costs.
The goal-oriented \Wording{planner} yields a slightly better result when no disturbance is \Wording{present} ($A=0$).
However, \Wording{whenever} noticeable disturbance is \Wording{present}, its performance degrades dramatically.
In addition, it halts before arriving at the goal because it \Wording{exceeds} the given time budget ($9h$ with $A \geq 0.8$).

The superior performance of our method can also be explained in the value function plots shown in Fig.~\ref{fig:value_funct}.
The classic MDP planner produces a discrete value function, i.e., the state-values are the same within \Wording{each $1km \times 1km$ cell}.
The resulting policy outputs the same actions \Wording{regardless of the position of the vehicle within each grid cell.}
\Wording{In contrast,}
\Wording{the proposed method outputs a continuous value function, which varies smoothly across continuous space, directly.} 
It can better characterize the actual optimal continuous value function.
Thus, the resulting policy is more flexible and more diverse, which can lead to more intelligent behavior. 

\RE{Finally, we perform detailed comparisons of approximate policy iteration using three different resolution parameters $h$ in Table~\ref{tab:discretization}.
When the resolution is set to $h=m$, a number of $(\frac{20}{m})^2$ evenly-spaced state points are generated.
In general, increasing the number of state points leads to better performance. 
We 
also observe that the gain between $h = 0.5km$ and $h = 1.0km$ is less than that between $h=1.0km$ and $h=2.0km$.
This implies that 
when the solution is close to the optimal, the gain in the performance by 
a finer discretization becomes small. 
By comparing with the classic PI planner, it can be seen that the performance of our method is only marginally affected after halving the resolution and remains a good performance equivalent or superior to the classic PI planner. 
}

\subsection{Evaluation with Ocean Data}

In addition to the gyre model, 
\Wording{we also used Regional Ocean Model System (ROMS)~\cite{shchepetkin2005regional} data to evaluate our approach.}
The dataset provides ocean current \Wording{forecasts} in the Southern California Bight region. 
\RE{Since ROMS only provides the ocean currents statistics at discrete locations, it cannot be directly used to evaluate the algorithms. 
To address this problem, we}
use Gaussian Process Regression (GPR) to interpolate ocean currents at every location on the 2-D ocean surface, \Wording{from a single time snapshot}.
We used $h=2km$ in our method when approximating the value function. 
For the classic MDP planner, the environment was discretized \Wording{into $28\times28$ grids} and each \Wording{cell} has a dimension of $2km \times 2km$. 
Fig.~\ref{fig:roms_traj} shows the trajectories \Wording{generated with} the three methods. Note that the maximum speed of the ocean currents from this dataset is around $3km/h$, which is similar to the vehicle's maximum linear speed. 
We can see that trajectories from approximate policy iteration are smoother (shorter distance and time costs) than those from the classic policy iteration approach. 

\vspace{-5pt}
\section{Conclusions}
In this paper, we propose a solution to solving the autonomous vehicle planning problem 
using a continuous approximate value function for the infinite horizon MDP and 
using finite element methods as key tools. 
\Wording{Our method leads to an accurate and continuous form of the value function even
if we only use a smaller number of discrete states and if only the first and second moments of the model transition probability are available.}
We achieve this by leveraging 
\Wording{a Taylor expansion of the Bellman equation to obtain a value function approximated by a diffusion-type partial differential equation,}
which can be naturally solved using finite element methods.
Extensive simulations and evaluations have demonstrated advantages of our methods for providing continuous value functions, and for better path results in terms of path smoothness, travel distance and time costs.

\appendix
\subsection{Appendix for Finite Element Methods}\label{appendix:fem}
We illustrate the ideas of the Galerkin approach in finite element methods through a diffusion equation.

Let $\Omega$ be a closed smoothly bounded domain in $\mathbb{R}^2$, and $\partial\Omega$ be its boundary that consists of two parts, i.e., $\partial\Omega = \partial\Omega_1 \cup \partial\Omega_2$. Let $b(x,y)$ be a scalar value, and let $q^T(x,y):=(q_1(x, y), q_2(x, y))$ be a vector. Assume that $D(x,y)$ is a 2-by-2 symmetric matrix satisfying $\sum D_{ij}\xi_i\xi_j \geq c\sum_{i}\xi^2_i$, and $c > 0$. Let $f(x,y)$ be a smooth function.

Consider the following 2-dimensional diffusion equation
\begin{align}\label{apd:diffusion-pde}
-\nabla\cdot D(x, y)\nabla v(x, y) + q^T(x,y)\,\nabla v(x,y) + \\ \nonumber 
b(x, y)\,v(x ,y) = f(x, y), 
\end{align}
with boundary conditions
\begin{subequations}\label{boundary-condition}
\begin{eqnarray}
   D(x,y)\,\nabla v(x ,y)\cdot \hat{\bm{n}} &=& 0, \mbox{  on } \partial\Omega_1 \label{boundary-condition-a}\\
    v(x, y) &=& \hat{u}(x,y), \mbox{  on } \partial\Omega_2, \label{boundary-condition-b}
\end{eqnarray}
\end{subequations}
where $\hat{\bm{n}}$ is the unit of outward normal to $\partial \Omega$, and $\hat{u}(x,y)$ is a given boundary value function.
The condition (\ref{boundary-condition-a}) is a type of homogeneous Neumann condition (treated as natural boundary condition \cite{hughes2012finite}), and (\ref{boundary-condition-b}) can be thought of as a Dirichlet condition (treated as essential boundary condition \cite{hughes2012finite}) in literature \cite{Evens2010}.

The Galerkin method consists of dividing the domain $\Omega$ into a finite number of elements, and then using the variational form (also referred to the weak form) of the partial differential equation to construct the solution through the linear combination of basis functions over elements.

The variational or weak form of the problem amounts to multiplying the both sides of Eq. (\ref{apd:diffusion-pde}) by a sufficiently smooth test function $\omega(x, y)$. By using integration-by-parts formula and 
boundary condition Eq. (\ref{boundary-condition}), we have
\begin{align}
    \int_{\Omega} \left(\nabla v\cdot D\nabla \omega + w\,q^T\,\nabla v + b\, v \,\omega\ - f\,\omega\right)dx\,dy = 0, 
    \label{weak-diffusion-pde}
\end{align}
where the test function $w(x,y)=0$ on $\partial \Omega_2$. It can be shown the solution to this variational form is also the solution to the original form. 

Then we partition the continuous domain $\Omega$
into suitable discrete elements. 
Typically triangle elements are applied because they are general enough to handle all kinds of domain geometry. Denote the mesh, the discretization of the domain consisting of triangle elements, by $\Omega^h$. Our aim is to construct an approximate solution of the form
\begin{align}
    v^h = \sum_{i=1}^N a_i \phi_i(x, y), 
    \label{approx-v}
\end{align}
where $\phi_i(x,y)$ are basis functions and $N$ depends on the nodes on elements to construct these basis functions (see, for example, Section 4 in the book \cite{oden2012introduction}). We use the Lagrange interpolation polynomials as basis functions, each of which is constructed based on vertices of the triangle element.

When we represent the test function by a linear combination of basis functions, i.e., $\omega(x, y)=\sum_{i=1}^n c_i\phi_i(x, y)$ and substitute it to Eq.~\eqref{weak-diffusion-pde}, we get an equation for the weights $c_i$ and $a_i$.
Because coefficients $c_i$ should be arbitrary, this along with condition (\ref{boundary-condition-b}) leads to a coupled system 
of linear algebraic equations
\begin{align}
    KA = F,
    \label{stiffen-eqn}
\end{align}
where $A=(a_1, \ldots, a_N)^T$.
Each entry of matrix $K$ corresponds to basis function involved integrals on the right-hand side of Eq. (\ref{weak-diffusion-pde}):
\begin{align}
    K_{i,j}=\int_{\Omega^h} \left(\nabla\phi_i\cdot D\nabla\phi_j + \phi_i\,q^T\,\nabla\phi_j + b\,\phi_i\,\phi_j\right)\,dx\,dy, 
    \label{stiffen-K}
\end{align}
and $F$ corresponds to the products with $f$: 
\begin{align}
    F_{i}=\int_{\Omega^h} f\,\phi_i(x,y)\,dx\,dy.
    \label{stiffen-F}
\end{align}
Solving this linear system gives the estimates of $a_i$, i.e., the approximate value function $v(s)$.

So far we do not explicitly explain in details how to deal with the boundary function $\hat{u}(x,y)$ in condition (\ref{boundary-condition-b}) in Eq.(\ref{stiffen-eqn}). The essential boundary condition Eq.(\ref{boundary-condition-b}) must be taken into account when we construct the approximation (\ref{approx-v}) and some modifications to $F_i$ may be done. We refer the readers to the aforementioned books for details.

{
\bibliographystyle{IEEEtran}
\bibliography{ref,finiteElementMethod,approximateMDP,HJB-control}
}
\end{document}